\documentclass[letterpaper, 10 pt, journal, twoside]{IEEEtran}

\IEEEoverridecommandlockouts                              

\usepackage[resetlabels]{multibib}
\usepackage{subfiles}
\usepackage{csquotes}
\usepackage{gensymb}
\usepackage{pifont}
\usepackage{textcomp}
\usepackage[table]{xcolor}
\usepackage{todonotes}
\usepackage{blindtext}
\usepackage{soul}
\usepackage{graphicx}
\usepackage{mathtools}
\usepackage{multirow}
\usepackage{flushend}
\usepackage{booktabs}
\usepackage{array}
\usepackage{amsmath} 
\usepackage{amssymb} 
\usepackage{amsfonts}
\usepackage{bm} 
\usepackage{siunitx}
\usepackage{cancel}
\usepackage{microtype}
\usepackage{xcolor}
\usepackage[breaklinks,colorlinks]{hyperref}
\usepackage{caption}
\usepackage{subcaption}
\usepackage{cite}
\usepackage[export]{adjustbox}
\usepackage{cuted}
\newcites{S}{References}

\usepackage[capitalize]{cleveref}
\crefname{section}{Sec.}{Secs.}
\Crefname{section}{Section}{Sections}
\Crefname{table}{Table}{Tables}
\crefname{table}{Tab.}{Tabs.}

\definecolor{mygreen}{HTML}{00A64F}
\definecolor{myred}{HTML}{ED1B23}

\newcommand{\secref}[1]{Sec.~\ref{#1}}
\renewcommand{\eqref}[1]{Eq.~(\ref{#1})}
\newcommand{\figref}[1]{Fig.~\ref{#1}}
\newcommand{\tabref}[1]{Tab.~\ref{#1}}

\usepackage{pgfplots}
\pgfplotsset{width=\linewidth,compat=1.9}
\definecolor{color_red}{RGB}{228,26,28}
\definecolor{color_blue}{RGB}{55,126,184}
\definecolor{color_green}{RGB}{77,175,74}
\definecolor{color_purple}{RGB}{152,78,163}
\definecolor{color_orange}{RGB}{255,127,0}
\definecolor{color_brown}{RGB}{166,86,40}
\definecolor{color_pink}{RGB}{247,129,191}
\usepackage{xcolor}
\usepackage{scalerel}

\newcommand{\rebuttal}[1]{{#1}}

\AfterEndEnvironment{strip}{\leavevmode}

\newcommand{\net}{INoD}

\title{\LARGE \bf
INoD: Injected Noise Discriminator for Self-Supervised Representation Learning in Agricultural Fields}

\author{Julia Hindel$^{1}$, Nikhil Gosala$^{1}$, Kevin Bregler$^{2}$, and Abhinav Valada$^{1}$
\thanks{$^{1}$University of Freiburg, $^{2}$Fraunhofer IPA}%
\thanks{This work was partly funded by the
German Research Foundation (DFG) Emmy Noether Program grant number 468878300.}
}

\begin{document}

\maketitle
\thispagestyle{empty}
\pagestyle{empty}

\begin{abstract}
Perception datasets for agriculture are limited both in quantity and diversity which hinders effective training of supervised learning approaches. Self-supervised learning techniques alleviate this problem, however, existing methods are not optimized for dense prediction tasks in agricultural domains which results in degraded performance.
In this work, we address this limitation with our proposed Injected Noise Discriminator~(\net) which exploits principles of feature replacement and dataset discrimination for self-supervised representation learning. \net~interleaves feature maps from two disjoint datasets during their convolutional encoding and predicts the dataset affiliation of the resultant feature map as a pretext task. 
Our approach enables the network to learn unequivocal representations of objects seen in one dataset while observing them in conjunction with similar features from the disjoint dataset. 
This allows the network to reason about higher-level semantics of the entailed objects, thus improving its performance on various downstream tasks. Additionally, we introduce the novel Fraunhofer Potato 2022 dataset consisting of over 16,800 images for object detection in potato fields. Extensive evaluations of our proposed \net~pretraining strategy for the tasks of object detection, semantic segmentation, and instance segmentation on the Sugar Beets 2016 and our potato dataset demonstrate that it achieves state-of-the-art performance.
\end{abstract}


\section{Introduction}
\label{sec:introduction}
In today's times, there is an ever-growing urgency to make agricultural practices ecologically sustainable while simultaneously improving farm throughput. Precision agriculture provides a solution to this challenge via the use of precise intervention techniques such as targeted spraying of chemicals, as well as using gripper arms, stomping feet, and lasers to destroy weeds while preserving crops. However, precision agriculture heavily relies on plant detection and segmentation which is often challenging due to their varied appearance throughout their growth cycle. This is further exacerbated by extensive overlaps with neighboring plants which results in their edges being indistinguishable from one another. Additionally, the lack of labeled datasets encompassing different crops hinders the development of fully-supervised plant detection and segmentation approaches. Self-supervised learning~(SSL) techniques help combat the outlined label deficiency by pretraining the network on pseudo labels obtained from self-derived pretext tasks before finetuning on the downstream target task. The pretraining step enables the network to learn the underlying semantics of the image which helps it better adapt to a wide range of relevant tasks and datasets. 

\begin{figure}
    \centering
    \includegraphics[width=0.9\linewidth]{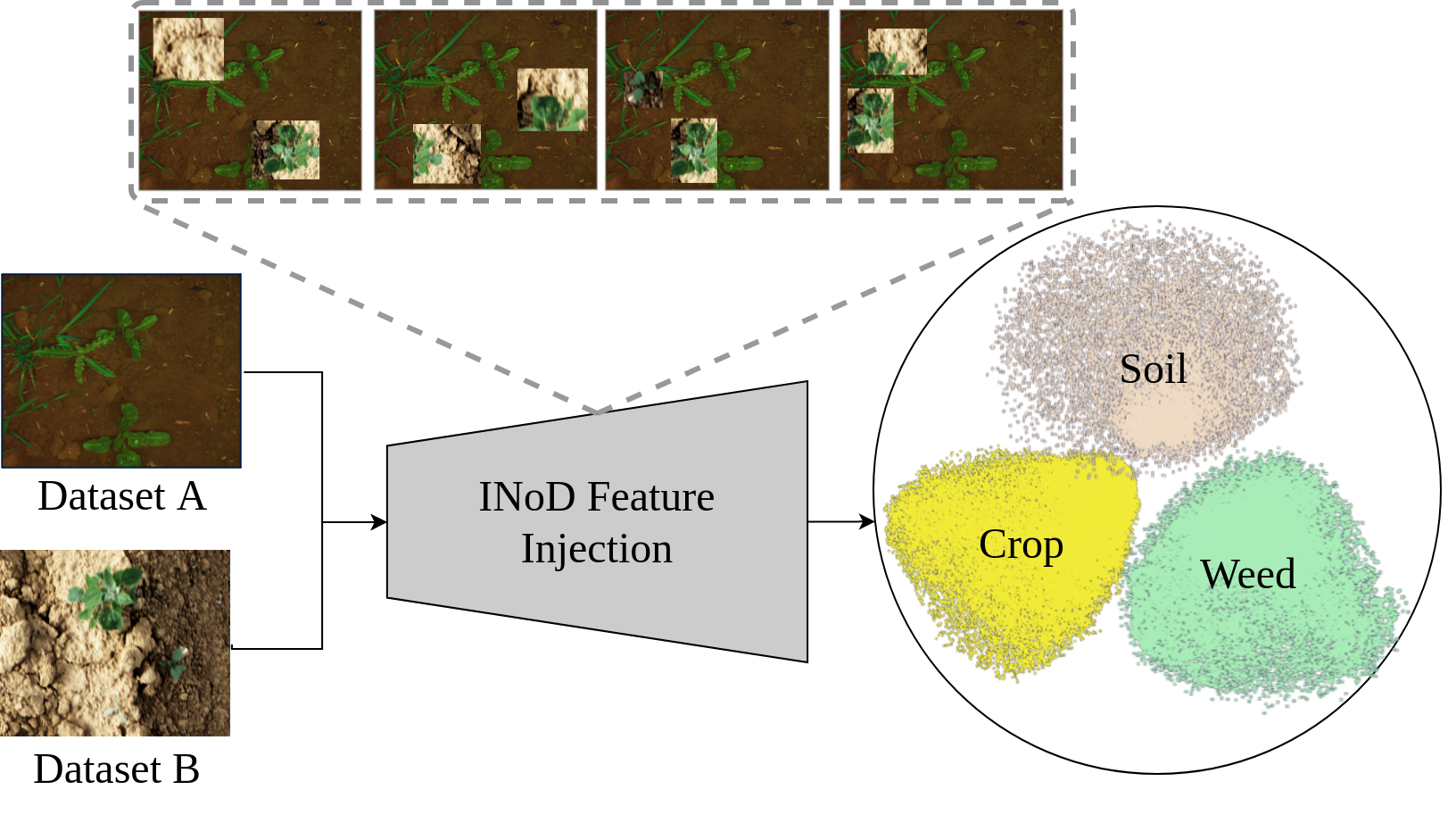}
    \caption{\net~interleaves convolutional features of two disjoint agricultural datasets during their convolutional encoding to learn semantically meaningful representations of entailed soil, crop, and weed in a self-supervised manner.}
    \label{fig:teaser}
    \vspace{-0.3cm}
\end{figure}

Existing SSL approaches in the agricultural domain rely on techniques such as self-labeling~\cite{cit:ndvi-paper}, \rebuttal{feature decorrelation~\cite{roggiolani2023icra-odsp}}, and contrastive learning~\cite{guldenring2021} for training their models in a label-efficient manner. However, self-labeling approaches rely on the NDVI metric~\cite{cit:ndvi-paper} which is often very susceptible to varying crop color and external lighting conditions. Further, existing agricultural models directly leverage \rebuttal{feature decorrelation and} contrastive learning techniques originally introduced for performing classification on the ImageNet dataset. Accordingly, these networks only focus on pretraining the network backbone which is often insufficient for dense prediction tasks such as segmentation and object detection as the task-specific heads are initialized randomly~\cite{hurtado2022semantic}. Moreover, little research has focused on adapting existing SSL techniques to the challenging domain of precision agriculture. Typical SSL approaches are largely tailored for the ImageNet dataset which significantly deviates from agricultural images, especially with regard to the scale, location, and frequency of objects.

To address the aforementioned limitations, we propose \textit{\textbf{I}njected \textbf{No}ise \textbf{D}iscriminator} \textit{(\net)}, a novel self-supervised representation learning approach for precision agriculture. The principle behind \net~is to interleave features of two disjoint datasets during their convolutional encoding phase and then determine the dataset affiliation of the resultant feature map using task-specific detection or segmentation heads. 
We hypothesize that \net~enables the network to learn unequivocal representations of various types of vegetation while observing them in conjunction with other valid features. This interleaving of features is performed at multiple levels during the forward pass of the network which prevents the trivial distinction of datasets based on generic dataset-specific statistics. Consequently, the network is constrained to determine the origin of features by reasoning about higher-level semantics of the entailed objects. This higher level reasoning enables the network to learn rich representations which improve its overall performance on downstream tasks. We also introduce the novel Fraunhofer Potato 2022 (FP22) dataset which is one of the first publicly available potato datasets with crop and weed annotations.
We perform extensive evaluations of \net~on the downstream tasks of object detection, semantic segmentation, and instance segmentation. Our experiments on the Sugar Beets 2016~\cite{Chebrolu2017AgriculturalRD} and the FP22 datasets demonstrate that \net~outperforms state-of-the-art SSL techniques by more than \rebuttal{\SI{1.3}{pp}} in terms of the AP and mIoU metrics. Additional ablation studies also demonstrate that our approach consistently exceeds the performance of standard SSL baselines across a wide range of finetuning splits and pretraining epochs.

Our main contributions can thus be summarized as follows:
\begin{enumerate}
  \item A novel SSL technique, \net, for pretraining any dense prediction network without further adaptations.
  \item A potato dataset comprising over $16,800$ images and $1,433$ object detection annotations.
  \item Several competitive SSL baselines for dense prediction tasks on agricultural datasets.
  \item Extensive evaluation and ablation of \net~on two agricultural datasets.
  \item Publicly available code and pretrained models at \\ \url{http://inod.cs.uni-freiburg.de}.
\end{enumerate}
\section{Related Work}
\label{sec:related-work}
In this section, we summarize existing works in SSL and present an overview of approaches that have been adapted to the domain of precision agriculture.

{\parskip=3pt
\noindent\textit{Self-Supervised Learning}: 
SSL is a type of unsupervised learning where pseudo labels are automatically derived from an unlabeled dataset~\cite{gosala2023skyeye}. Since the scope of SSL is wide-reaching, we limit this section to self-supervised pretraining approaches that are commonly employed in conjunction with downstream tasks such as image classification, object detection, and semantic segmentation.}

Early SSL approaches employed simple augmentation tasks such as predicting the transformation of images~\cite{Gidaris2018} or determining the spatial arrangement of image patches~\cite{Noroozi2016} for pretraining the network backbone. These approaches often rely on ad-hoc heuristics during pretraining which often fail to generalize over a wide variety of downstream tasks.
Recently, discriminative pretext tasks including contrastive and self-training methods have gained increased attention in the field of SSL. The pioneering contrastive models SimCLR~\cite{chen2020simCLR} and MoCo~\cite{He2020MoCo} have been experimentally shown to outperform supervised pretraining on various downstream tasks. Another direction of research omits the need for negative samples in contrastive learning by enforcing the proximity of positive pairs in the latent space. This proximity of positive samples is often realized using different variants of clustering~\cite{caron2021SwaV}, self-distillation~\cite{grill2020byol, caron2021DINO}, \rebuttal{feature decorrelation}~\cite{Zbontar2021BarlowTS} or Siamese networks~\cite{chen2020SimSiam}. 
In the context of object detection and instance segmentation, where the localization ability of the network is crucial~\cite{lang2022robust, mohan2022amodal}, a mismatch between the pretext and final task can easily occur when significantly different objectives are optimized in the different training phases.
Nevertheless, existing classification-based pretraining approaches, that learn translation and scale invariance, have been used for object detection resulting in low overall performance~\cite{Yang2021InstanceLF, Xiao2021RegionSR}.
Other works encourage the network to reason about localization using a dense contrastive loss function that only compares region-specific features obtained from known region-based correspondences~\cite{Pinheiro2020VADeR,Xiao2021RegionSR, Ding2022DeeplyUP}. Further extensions contrast global features in addition to region-specific features to generate robust representations~\cite{Wang2021DenseCL, Xie2021DetCoUC, lang2023self}. Parallel works use pre-processing steps such as selective search and unsupervised perceptual grouping to generate local object regions before applying contrastive learning~\cite{Henaff2021EfficientVP, Wei2021AligningPF}.

More recently, cut-and-paste pretraining strategies, wherein random crops of an image are pasted onto different backgrounds and then classified as foreground and background have been employed to learn rich representations for dense prediction tasks~\cite{Yang2021InstanceLF, Wang2022CP22}. 
These contrastive learning-based approaches often demand significant resources during pretraining due to their reliance on large batch sizes or queues of data samples. In this paper, we propose a novel SSL approach, \net, where feature maps from two disjoint datasets are interleaved during their convolutional encoding, and a network is then tasked to determine the dataset affiliation of the composite feature map. Our approach does not exhibit intra-batch dependencies and can thus be easily trained on a single GPU.
Moreover, these contrastive learning-based approaches need to be specifically adapted to pretrain the complete target network and often depend on finding the correct combination of loss terms. In contrast, our approach can be used with any network architecture without additional modifications and can directly be employed for various downstream dense prediction tasks. 

{\parskip=3pt
\noindent\textit{Self-Supervised Learning in Agriculture}: 
There are only a handful of SSL approaches in the agricultural domain. Zapata~\textit{et~al.}~\cite{Zapata2021SelfsupervisedFE} train a triple Siamese network to distinguish between plant seedlings and demonstrate its effectiveness using the image retrieval task.
\rebuttal{Besides, \cite{Choi2021SelfsupervisedRL} employs randomized color channel recognition to pretrain a network for fruit anomaly detection, while \cite{roggiolani2023icra-odsp} uses BarlowTwins~\cite{Zbontar2021BarlowTS} coupled with domain-adapted augmentations for leaf and plant segmentation.} 
SwAV~\cite{caron2021SwaV} is an SSL approach that was used to pretrain classification and segmentation networks on datasets of grassland, aerial farmland, and weed species~\cite{guldenring2021}. However, the authors report no significant improvement when using SSL \rebuttal{for} plant segmentation, thus highlighting the need for further research on leveraging SSL for dense prediction tasks in the agricultural domain.
Our proposed self-supervised \net~approach demonstrates exceptional performance for a range of dense prediction tasks such as object detection, semantic segmentation, and instance segmentation.}

\section{Technical Approach}
\label{sec:technical-approach}

\begin{figure*}
    \centering
    \includegraphics[width=\linewidth]{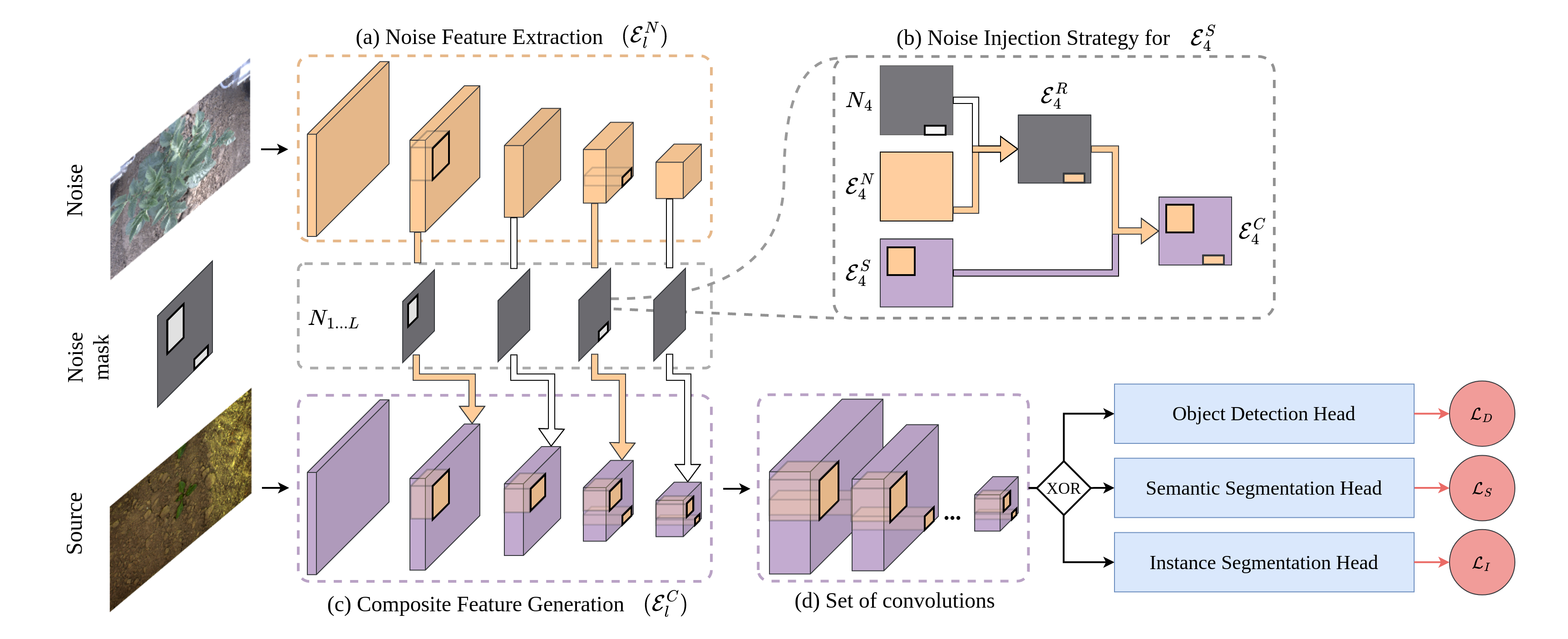}
    \caption{INoD injects features from a disjoint noise dataset into various feature levels of a source dataset during their convolutional encoding. The noise injection is defined in binary layer-specific noise masks which originate from a randomly generated noise mask. The network is then trained end-to-end to determine the dataset affiliation with an object detection, semantic segmentation or instance segmentation head. In this figure, the ``set of convolutions'' refers to an FPN for the instance segmentation and object detection tasks, and a DPC module for semantic segmentation.}
    \label{fig:overview-inod}
    \vspace{-0.3cm}
\end{figure*}

In this section, we first present an overview of the proposed self-supervised \net~pretraining approach which is tailored for dense prediction tasks in the agricultural domain. The goal of injected feature discrimination is to make the network learn unequivocal representations of objects from one dataset while observing them alongside objects from a disjoint dataset. Accordingly, the pretraining is based on the premise of injecting features from a disjoint \textit{noise} dataset into different feature levels of the original \textit{source} dataset during its convolutional encoding. The network is then trained to determine the dataset affiliation of the resultant feature map which enables learning rich discriminative feature representations for different objects in the image. \figref{fig:overview-inod} provides an overview of our approach.

\subsection{Overview of \net} \label{subsec:intro}
Our approach provides a \textit{drop-in} solution to pretrain any network using our novel injected feature discrimination strategy.
\net~encourages the network to learn semantically representative features, thus improving the transfer learning ability for dense prediction tasks. 
This can be attributed to the fact that \net~forces the network to decipher feature origin based on a higher level semantic understanding of the feature maps instead of generic dataset-specific statistics. Our approach comprises four phases, namely, (i) multi-level noise feature generation, (ii) random noise mask generation, (iii) iterative noise injection and subsequent forward pass of the multi-level source image features, and (iv) dense prediction tasks to determine the origin of features in the composite feature map.

First, we use an off-the-shelf network backbone $\Theta$ to compute multi-scale feature maps for the noise image: $\Theta(\mathcal{I}_N) \rightarrow \{ \mathcal{E}^N_{1}, \mathcal{E}^N_{2}, ..., \mathcal{E}^N_{L}\}$ (\figref{fig:overview-inod}(a)). Second, we generate random layer-specific binary noise masks, $N_{l}$, for each scale $l$ to determine the spatial locations at which the noise features are injected into the source features.
Third, we extract region-specific noise features, $\mathcal{E}^R_l$, by multiplying each of the multi-scale noise features, $\mathcal{E}^N_{l}$, with their corresponding layer-specific noise mask $N_l$. Then, we iteratively inject $\mathcal{E}^R_l$ at equivalent locations in the source feature map $\mathcal{E}^S_{l}$ to generate a composite feature map $\mathcal{E}^C_{l} = f(\mathcal{E}^N_{l}, \mathcal{E}^S_{l}, N_{l})$ (\figref{fig:overview-inod}(b)). We then perform the subsequent source network traversal step of $\Theta$ to generate the next level feature map $\mathcal{E}^S_{l+1}$.
Thus, noise injected in early feature maps is carried through higher layers of the network backbone as shown in \figref{fig:overview-inod}(c). \rebuttal{We intertwine features only along the height and width dimensions of the feature map to ensure the integrity of convolutional feature representations in the channel dimension.} Finally, we pass the composite feature maps through a set of convolutional layers to further entangle features and generate coherent multi-scale representations (\figref{fig:overview-inod}(d)). We then provide these feature maps as input to downstream task heads such as object detection, semantic segmentation, \rebuttal{or} instance segmentation to infer the dataset affiliation of the composite feature maps. 

\subsection{Noise Mask Generation}
\label{subsec:noise-mask-generation}

Noise mask generation forms one of the core components of the noise injection protocol and defines the regions where source features are replaced by noise features. A noise mask $N$ is primarily defined by its granularity which determines the smallest possible volume that can be replaced in a feature map. In other words, a noise mask with high granularity allows for the substitution of small feature regions and vice versa. In this paper, we select one feature scale from the network backbone as our reference size and correspondingly specify the minimal feature replacement dimensions according to its effective receptive field. Once the granularity is defined, we can trivially generate the noise mask by randomly sampling noise injection positions on a blank canvas.
However, these masks often exhibit disjoint and similarly-shaped noise injection positions which reduces the training diversity and prevents the network from being able to predict diverse shapes and sizes. We mitigate this problem by first generating random binary sample masks of size $3\times3$ with probability $P(1)=\frac{2}{3}$, rescaling them to $[\frac{2}{3}, \frac{1}{6} \times$ reference scale dimensions], and then randomly placing them within the noise mask bounds. This strategy allows us to create highly complex noise patterns that facilitate learning semantically meaningful representations for all three dense prediction tasks.

\subsection{Noise Injection}
\label{subsec:noise-injection}
Our noise injection protocol comprises two stages, namely, (i)~layer-specific noise mask extraction, and (ii)~layer-specific noise injection. First, we generate layer-specific noise masks $N_{1...L}$ by randomly sub-dividing the noise mask $N$, defined in \secref{subsec:noise-mask-generation}, such that non-zero regions in $N$ are sampled only once in $N_{1 ... L}$ to ensure that $N=\Sigma_{l}N_l$ \rebuttal{as visualized in \figref{fig:noise-mask}}. We then estimate the region-specific noise features for layer $l$, $\mathcal{E}^R_{l}$, by multiplying $N_l$ with $\mathcal{E}^N_l$. Finally, we inject noise into a source feature level by replacing regions of $\mathcal{E}^S_{l}$ with $\mathcal{E}^N_{l}$ at equivalent locations. Mathematically, 
\begin{equation}
\mathcal{E}^C_{l} = \mathcal{E}^N_{l} \cdot N_{l} + \mathcal{E}^S_{l} \cdot \neg N_{l}.
\end{equation}
The next source feature level is then computed as
\begin{equation}
\mathcal{E}^S_{l+1} = \text{conv}_{l+1}(\mathcal{E}^C_l).
\end{equation}

\subsection{Task-Specific Ground Truth Label Generation} \label{subsec:pseudo-label}

We supervise the dataset affiliation of the composite feature map by generating task-specific pseudo labels \rebuttal{as shown in \figref{fig:labels}}. The following sections outline our pseudo label generation process \rebuttal{for three dense prediction tasks}.
 
{\parskip=3pt
\noindent\textit{Object Detection}: \label{OD_head}
We supervise the object detection pretext task by creating bounding box pseudo labels from the noise mask \rebuttal{$N$} generated in \secref{subsec:noise-mask-generation}. \rebuttal{First, we compute} contours around non-zero elements in $N$ and then \rebuttal{draw} tight axis-aligned bounding boxes around them. For the specific case when noise mask elements only share a common corner, we avoid computing a common contour and instead compute element-specific contours to prevent a large performance drop caused by sparse bounding boxes.
Drawing boxes around contours instead of around every independent noise element allows us to generate diverse bounding box pseudo labels in terms of both box size and scale which improves the model performance.}

{\parskip=3pt
\noindent\textit{Semantic Segmentation}: \label{SS_head}
For the semantic segmentation task, we create the corresponding pseudo label by resizing the noise mask $N$ to the pre-determined output resolution using the nearest neighbor interpolation algorithm.}

{\parskip=3pt
\noindent\textit{Instance Segmentation}: 
We compute instance segmentation pseudo labels by first generating bounding box pseudo labels and then post-processing them to extract instance IDs of different elements in $N$. The post-processing step entails \rebuttal{extracting} non-zero \rebuttal{pixels} within a bounding box and assigning \rebuttal{them a} common instance label. We follow the aforementioned two-step process to ensure consistency between the bounding box and instance segmentation pseudo labels which are needed for pre-training the Mask R-CNN architecture.}

\subsection{Loss functions} \label{subsec:loss}
We pretrain \rebuttal{networks} for \rebuttal{target} prediction tasks using \rebuttal{only the} corresponding task-specific losses. In this paper, we pretrain \rebuttal{a} semantic segmentation network by minimizing the binary focal loss ($\mathcal{L}_S$)~\cite{Lin2017FocalLF}, optimize \rebuttal{an} object detection network using the standard detection-specific loss components of the Faster R-CNN architecture ($\mathcal{L}_D$), and train \rebuttal{an} instance head using Mask R-CNN losses such as proposal, class, bounding box, and mask loss ($\mathcal{L}_I$)~\cite{He2017maskrcnn}.
We further describe the network architectures employed in our experiments in \secref{subsec:experimental-settings}.
\section{Experimental Evaluation}
In this section, we quantitatively and qualitatively evaluate the performance of \net~on three dense prediction tasks and also provide a comprehensive ablation study to demonstrate the importance of our contribution. We first present an overview of the \rebuttal{used} agricultural datasets and then describe the experimental settings for the pretraining and finetuning pipelines.

\subsection{Datasets}
\parskip=3pt
\noindent\textit{Sugar Beets 2016 (SB16)}: This dataset comprises recordings of the two months farming cycle of a sugar beet field near Bonn, Germany~\cite{Chebrolu2017AgriculturalRD}. \rebuttal{It} consists of $123,062$ samples while ground truth semantic labels are provided for a subset of $12,196$ images. We generate instance ground truth labels for these samples by drawing contours around the semantic labels using morphological closure and vegetation heuristics. For pretraining, we randomly divide the dataset into splits of $110,756$ training and $12,306$ validation samples. For the finetuning step, we divide the labeled samples into train, validation, and test splits based on their timestamp. Specifically, we finetune all models using the first $9,137$ labeled samples, validate them on the next $612$ images, and present the results on the remaining $2,447$ samples. We split the dataset using their timestamps to prevent the networks from remembering the characteristics of vegetation on different farming days, and instead allow them to demonstrate their true generalization capabilities. 

\begin{figure}
    \centering
    \begin{subfigure}[b]{0.4\columnwidth}
   	\includegraphics[width=\linewidth, frame]{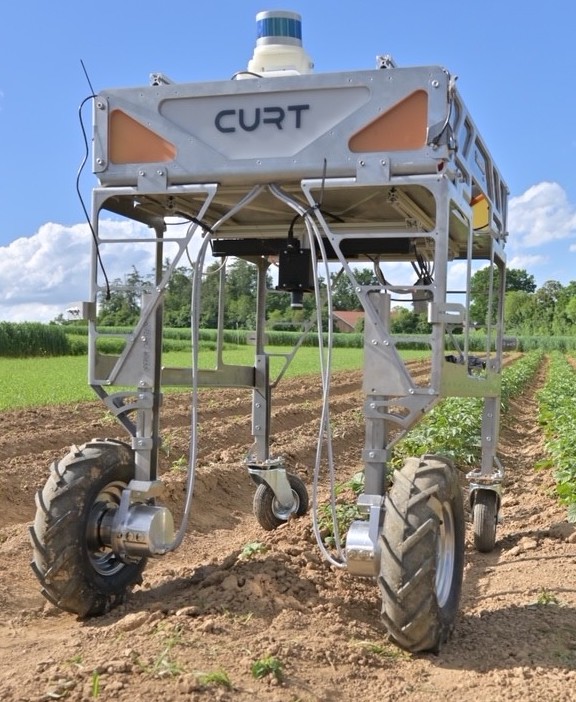}
  \end{subfigure}
  \begin{subfigure}[b]{0.322\columnwidth}
   	\includegraphics[width=\linewidth, frame]{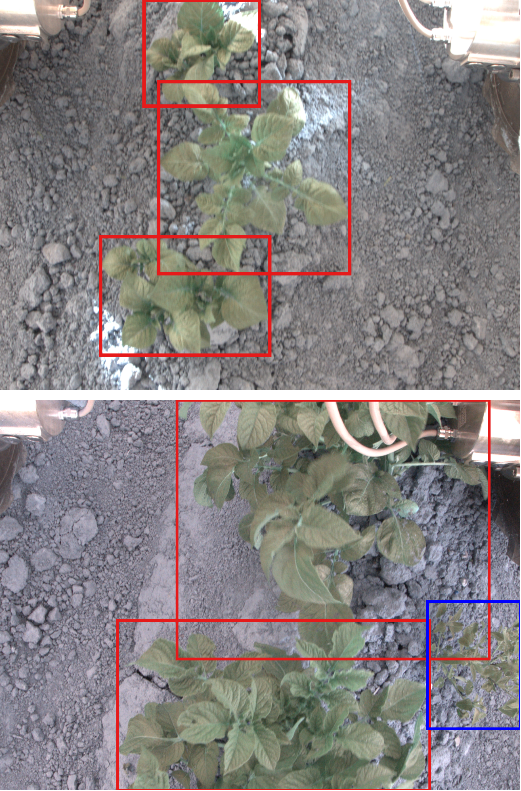}  
\end{subfigure}
\caption{Agricultural robot by Fraunhofer IPA (left) used to collect our FP22 dataset and samples from our FP22 potato dataset (right).}
\label{fig:curt}
\vspace{-0.3cm}
\end{figure}

\parskip=3pt
\noindent\textit{Fraunhofer Potato 2022 (FP22)}: This is a recent dataset by Fraunhofer IPA which was recorded using an agricultural robot at a potato cultivation facility in the outskirts of Stuttgart, Germany. The robot depicted in \figref{fig:curt} comprises a Jai Fusion FS 3200D 10G camera mounted at the bottom of the robot chassis at a height of \SI{0.8}{\meter} above the ground. The dataset contains $16,891$ images obtained from two different stages in the farming cycle of which a subset of $1,433$ images have been annotated with bounding box labels following a peer-reviewed process. For the pretraining step, we train the model using $15,202$ training samples and validate it on the remaining $1,689$ samples. Similar to SB16, we divide the labeled samples into three splits based on their timestamp yielding $867$ train, $169$ validation, and $397$ test samples. We make this dataset publicly available with this work at \url{http://inod.cs.uni-freiburg.de}.

\subsection{Experimental Setup}
\label{subsec:experimental-settings}

We evaluate our self-supervised \net~pretraining strategy on the dense prediction tasks of instance segmentation, semantic segmentation, and object detection. Accordingly, we employ a Mask R-CNN network~\cite{He2017maskrcnn} for instance segmentation, a ResNet-50 backbone with a DPC-based head~\cite{dpchead2018} for semantic segmentation, and a Faster R-CNN (R50+FPN) model~\cite{ren2016fasterrcnn} for object detection. The following sections detail the various parameters used in the pretraining and finetuning stages.

\subsubsection{Self-supervised Pretraining}\label{subsec:SSL-setting}

We perform self-supervised pretraining for dense prediction tasks using image crops of size $224 \times 224$ and perturb these images using random combinations of horizontal flips, gaussian blur, random grayscale, and color jitter as proposed in MoCo-v2~\cite{chen2020mocov2}. We \rebuttal{randomly initialize the model weights and update them }using the SGD optimizer with batches of size $256$, momentum of $0.9$, and weight decay of $0.0001$. We adopt a learning rate of $0.02$ which we decay by a factor of $0.1$ after $60\%$ and $80\%$ of the \rebuttal{training steps}. We train the models for $100$ and $200$ epochs on the SB16 and FP22 datasets respectively and ablate over different pre-training lengths for FP22 in~\secref{ablation:epochs}. We inject noise features into the outputs of all residual blocks of the ResNet-50 backbone. For instance segmentation, we inject noise with a granularity of $1/32$, while we employ noise masks with a granularity $1/4$ for semantic segmentation and object detection. Further, we limit the quantity of noise to 20\% of the final composite feature map for object detection and instance segmentation, while we apply 40\% of noise features for semantic segmentation. We ablate different granularities and quantities of noise injection in~\secref{ablation:granularity} and \secref{ablation:quantity}, respectively.
\rebuttal{We} utilize the \rebuttal{FP22 dataset} as noise for the SB16 dataset and \rebuttal{vice versa}. Lastly, we normalize both the source and noise images using the dataset statistics computed using the source dataset.
 
\subsubsection{Supervised Finetuning}

We first prepare the task-specific pretrained networks for finetuning by recomputing the batch normalization parameters for $400$ iterations following the approach outlined in~\cite{Wu2021RethinkingI}. This pre-processing step negates any discrepancy between pretraining and finetuning dataset distributions and creates a fair starting point for model finetuning. We then finetune these networks on images of size $800 \times 800$ for $20$ epochs on the SB16 dataset and $50$ epochs on the FP22 dataset. We optimize our model using SGD with a batch size of $16$, base learning rate of $0.02$, momentum of $0.9$, and weight decay of $0.0001$. Similar to pretraining, we decay the learning rate by a factor of $0.1$ after $60$\% and $80$\% of training epochs. We augment the train split using random horizontal flips for instance segmentation and object detection, while we also incorporate random gaussian blur and color jitter for the semantic segmentation task.

\subsection{Quantitative Results}
\label{subsec:quant-results}
We benchmark \net~against \rebuttal{six} popular pretraining baselines, namely, ImageNet pretraining, MoCo-v2~\cite{chen2020mocov2}, BYOL~\cite{Wang2021DenseCL}, InsLoc~\cite{Yang2021InstanceLF}, DenseCL~\cite{Wang2021DenseCL} \rebuttal{and domain-specific BarlowTwins (AgriBT)~\cite{roggiolani2023icra-odsp}}. We ensure fair comparison between all the baselines by following the same experimental settings outlined in \secref{subsec:experimental-settings} and using their published codebases wherever possible. \tabref{tab:maskrcnn} and \tabref{tab:semseg} present the results for finetuning the network on the SB16 dataset for instance and semantic segmentation, while \tabref{tab:fastrcnn} presents the finetuning results for object detection on the FP22 dataset. \rebuttal{We evaluate the models using the COCO evaluation metrics, and thereby compute mAP, AP\textsubscript{crop} and AP\textsubscript{weed} over IoU=$50$:$95$}.

\begin{table}
\footnotesize
\centering
\caption{Evaluation of instance segmentation on Sugar Beets 2016. All metrics are reported in [$\%$] and averaged over three runs.
}
\label{tab:maskrcnn}
\setlength\tabcolsep{3.7pt}
 \begin{tabular}{l|cc|ccc}
 \toprule
 \textbf{Pretraining} & \textbf{AP\textsubscript{crop}} & \textbf{AP\textsubscript{weed}} &\textbf{mAP} & \textbf{AP\textsubscript{75}} & \textbf{AP\textsubscript{50}} \\
 \midrule
Supervised~(IN) & 38.85 & 12.58 & 25.72 & 18.84 & 51.28  \\
MoCo-v2~\cite{chen2020mocov2} & 42.68 & 15.12 & 28.90 & 22.76 & 55.33 \\
BYOL~\cite{grill2020byol} & 48.15 & 14.04 & 31.10 & 31.81 & 55.49 \\
InsLoc~\cite{Yang2021InstanceLF} & 42.58 & 12.76 & 27.67 & 23.30 & 51.62 \\
DenseCL~\cite{Wang2021DenseCL} & 44.94 & 15.34 & 30.14 & 27.65 & 55.45 \\
\rebuttal{AgriBT ~\cite{roggiolani2023icra-odsp}} & \rebuttal{42.87} & \rebuttal{13.83} & \rebuttal{28.35} & \rebuttal{24.95} & \rebuttal{50.66} \\
\cmidrule{1-6}
\net~(Ours) & \rebuttal{\textbf{54.96}} & \rebuttal{\textbf{15.48}} & \rebuttal{\textbf{35.22}} & \rebuttal{\textbf{41.35}} & \rebuttal{\textbf{58.30}} \\
\bottomrule
\end{tabular}
\vspace{-0.2cm}
\end{table}

We observe from \tabref{tab:maskrcnn} that our approach outperforms all the pretraining baselines by more than \rebuttal{\SI{4.12}{pp}} on the instance segmentation task. This substantial improvement in performance is a consequence of a distinct increase in both the crop and weed classes and can be attributed to the rich semantic understanding brought about by our pretraining framework. Further, we observe that \net~significantly outperforms \rebuttal{the baselines by \SI{9.54}{pp}} on the AP\textsubscript{75} metric \rebuttal{while surpassing them by only \SI{2.97}{pp} on AP\textsubscript{50}}. This disparity between the AP\textsubscript{50} and AP\textsubscript{75} metrics highlights the uncertainty of segmentation predictions of the baselines in challenging regions such as leaf structures and leaf boundaries. We also visually verify this observation using qualitative results in \secref{qual}.

\begin{table}
\footnotesize
\centering
\caption{Evaluation of semantic segmentation on Sugar Beets 2016. All metrics are reported in [$\%$] and averaged over three runs.}
\label{tab:semseg}
\setlength\tabcolsep{3.7pt}
 \begin{tabular}{l|ccc|c}
 \toprule
 \textbf{Pretraining} & \textbf{IoU\textsubscript{crop}} & \textbf{IoU\textsubscript{weed}} & \textbf{IoU\textsubscript{soil}} &
 \textbf{mIoU} \\
 \midrule
Supervised~(IN) & 31.51 & 9.57 & 98.97 & 46.35 \\
MoCo-v2~\cite{chen2020mocov2} & 34.92 & 8.06 & 97.44 & 46.81 \\
BYOL~\cite{grill2020byol} & 23.01 & 7.71 & 97.46 & 42.73 \\
InsLoc~\cite{Yang2021InstanceLF} & 18.40 & 9.09 & 98.95 & 41.81 \\
DenseCL~\cite{Wang2021DenseCL} & 30.21 & 7.98 & \textbf{99.16} & 45.24 \\
\rebuttal{AgriBT}~\cite{roggiolani2023icra-odsp} & \rebuttal{25.84} & \rebuttal{\textbf{10.23}} & \rebuttal{96.97} & \rebuttal{44.35}\\
\midrule
\net~(Ours) & \rebuttal{\textbf{36.66}} & \rebuttal{9.98} & \rebuttal{97.74} & \rebuttal{\textbf{48.13}} \\
\bottomrule
\end{tabular}
\vspace{-0.3cm}
\end{table}

\tabref{tab:semseg} presents the performance of \net~for semantic segmentation in terms of mIoU. The mIoU \rebuttal{metric} is computed as the average over the crop, weed, and background (soil) classes. We observe that \net~exceeds the best-performing baseline by \rebuttal{\SI{1.32}{pp}} which can be attributed to an improved segmentation ability on \rebuttal{the crop class. While AgriBT shows superior results on the weed class, it performs poorly on crop detection.} Further, we note that supervised ImageNet pretraining demonstrates superior results compared to three self-supervised baselines namely BYOL, InsLoc, and DenseCL for this task. This large disparity underlines the degraded performance of existing SSL algorithms for semantic segmentation in agriculture fields.

\begin{table}[t]
\footnotesize
\centering
\caption{Evaluation of object detection on Fraunhofer Potato 2022 dataset. All metrics are reported in [$\%$] and averaged over three runs.}
\label{tab:fastrcnn}
\setlength\tabcolsep{3.7pt}
 \begin{tabular}{l|cc|c}
 \toprule
\textbf{Pretraining} & \textbf{AP\textsubscript{crop}} & \textbf{AP\textsubscript{weed}} & \textbf{mAP} \\
 \midrule
Supervised (IN)  & 56.77 & \textbf{35.07} & 45.92 \\
MoCo-v2~\cite{chen2020mocov2}  & 58.05 & 32.39 & 45.22 \\
BYOL~\cite{grill2020byol}  & 56.46 & 34.04 & 45.25 \\
InsLoc~\cite{Yang2021InstanceLF}  & 58.31 & 30.40 & 44.36 \\
DenseCL~\cite{Wang2021DenseCL}  & 57.63 & 32.54 & 45.08 \\
\rebuttal{AgriBT ~\cite{roggiolani2023icra-odsp}} & \rebuttal{57.70} & \rebuttal{33.64} & \rebuttal{45.67} \\
\cmidrule{1-4}
\net~(Ours) & \textbf{60.85} & 33.24 & \textbf{47.05} \\
\bottomrule
\end{tabular}
\end{table}

\tabref{tab:fastrcnn} shows the results of \net~for object detection on our FP22 dataset. We observe that our approach outperforms the best SSL benchmark by \rebuttal{\SI{1.38}{pp}} and the supervised ImageNet pretraining by \SI{1.13}{pp}. This substantial improvement in the mAP score can be attributed to the better detection performance on the crop class. Further, we note that supervised ImageNet pretraining achieves the best AP score on the weed class, exceeding the closest self-supervised benchmark by \SI{1.03}{pp}. We highlight that ImageNet pretraining has an unfair advantage due to two main reasons, namely, (i) a significantly longer training schedule and (ii) a diverse pretraining dataset. ImageNet pretraining is trained for nearly three magnitudes longer as compared to SSL baselines (1.2~billion iterations for ImageNet vs 3~million iterations for SSL) and is also trained on a very diverse dataset which allows it to extract rich information from a wide variety of scenarios. Further, the relatively high performance of the ImageNet pretrained weights on the weed class can be reasoned with the fact that ImageNet comprises several instances of monocots (e.g. grass) which constitute the majority of observed weeds in our FP22 dataset. Nevertheless, our SSL framework is still able to extract useful information pertaining to the crop and weed classes allowing it to achieve competitive detection performance for both classes.

\subsection{Ablation Study} 
\label{ablation}
In this section, we study the impact of various architectural components and hyperparameters on the performance of our approach. We \rebuttal{execute} all ablation experiments on the object detection task using our FP22 dataset.

\subsubsection{Impact of Pretraining}
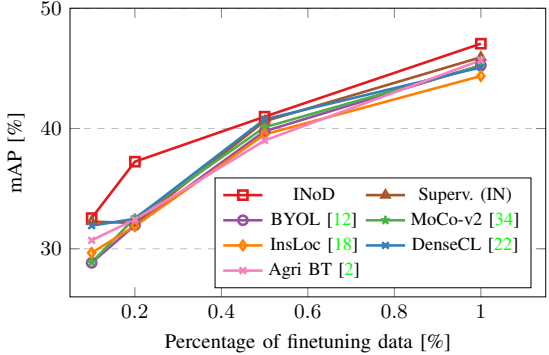
\begin{figure} 
        \centering
        \resizebox{0.82\columnwidth}{!}{%
        \begin{tikzpicture} [font=\small]
        \begin{axis}[
            title={},
            ylabel={mAP [\%]},
            legend style={font=\footnotesize},
            xlabel={Percentage of finetuning data [\%]},
            xmin=0.05, xmax=1.15,
            ymin=26, ymax=50,
            ytick={20,30,40,50},
            xtick={0,0.2,0.4,0.6,0.8,1},
            legend pos=south east,
            legend columns=2, 
            ymajorgrids=true,
            grid style=dashed,
            height=6cm
        ]

        \addplot[
            color=color_red,
            mark=square,
            line width=0.4mm,
            ]
            coordinates {
            (0.1,32.55)(0.2,37.24)(0.5,40.98)(1,47.05)
            };
        \addlegendentry{\net}
        \addplot[
            color=color_brown,
            mark=triangle,
            line width=0.4mm,
            ]
            coordinates {
            (0.1,32.26)(0.2,32.14)(0.5,40.60)(1,45.92)
            };
        \addlegendentry{Superv. (IN)}
        \addplot[
            color=color_purple,
            mark=o,
            line width=0.4mm,
            ]
            coordinates {
            (0.1,28.83)(0.2,31.95)(0.5,39.77)(1,45.25)
            };
        \addlegendentry{BYOL~\cite{grill2020byol}}
        \addplot[
            color=color_green,
            mark=star,
            line width=0.4mm,
            ]
            coordinates {
            (0.1,28.87)(0.2,32.57)(0.5,40.10)(1,45.22)
            };
        \addlegendentry{MoCo-v2~\cite{chen2020mocov2}}
                 \addplot[
            color=color_orange,
            mark=diamond,
            line width=0.4mm,
            ]
            coordinates {
            (0.1,29.66)(0.2,31.89)(0.5,39.51)(1,44.36)
            };
        \addlegendentry{InsLoc~\cite{Yang2021InstanceLF}}
        \addplot[
            color=color_blue,
            mark=x,
            line width=0.4mm,
            ]
            coordinates {
            (0.1,31.92)(0.2,32.49)(0.5,40.76)(1,45.08)
            };
        \addlegendentry{DenseCL~\cite{Wang2021DenseCL}}
        \addplot[
            color=color_pink,
            mark=x,
            line width=0.4mm,
            ]
            coordinates {
            (0.1,30.7)(0.2,32.43)(0.5,39.0)(1,45.67)
            };
        \addlegendentry{Agri BT~\cite{roggiolani2023icra-odsp}}
    \end{axis}
    \end{tikzpicture}}
\caption{Ablation study on the impact of finetuning with an increasing amount of data. The results are averaged over three runs.}
\label{fig:ablation-gt}
\vspace{-0.3cm}
\end{figure}
In this study, we analyze the impact of pretraining on the overall performance of the model by varying the amount of labeled data during the finetuning step. Therefore, we reduce the amount of labeled data during finetuning to $10$\%, $20$\%, and $50$\% of the original finetuning dataset size but maintain a constant number of iterations during finetuning. \figref{fig:ablation-gt} shows the mAP scores for the different percentage splits. We observe that our model consistently outperforms all the baselines even when using a small number of labeled samples during finetuning, thus highlighting the benefit of our \net~strategy. Our model also shows superior performance when using only $20$\% of finetuning samples, outperforming all the baselines by more than \SI{4.67}{pp}.

\subsubsection{Length of Pretraining}\label{ablation:epochs}
\begin{figure}
\centering
        \resizebox{0.8\columnwidth}{!}{%
        \begin{tikzpicture} [font=\small]
        \begin{axis}[
            title={},
            ylabel={mAP [\%]},
            xlabel={Number of pretraining epochs},
            xmin=80, xmax=820,
            ymin=42.5, ymax=48.5,
            ytick={44, 46, 48},
            xtick={100, 200, 400, 800},
            legend pos=south east,
            legend columns=2, 
            legend style={font=\footnotesize},
            ymajorgrids=true,
            grid style=dashed,
            height=6cm
        ]

        \addplot[
            color=color_red,
            mark=square,
            line width=0.4mm,
            ]
            coordinates {
            (100, 45.95)(200, 47.05)(400, 47.25)(800, 47.87)
            };
        \addlegendentry{\net}
        \addplot[
            color=color_green,
            mark=star,
            line width=0.4mm,
            ]
            coordinates {
            (100, 46.05)(200, 45.22)(400, 47.02)(800, 47.34)
            };
        \addlegendentry{MoCo-v2~\cite{chen2020mocov2}}
        \addplot[
            color=color_purple,
            mark=o,
            line width=0.4mm,
            ]
            coordinates {
            (100, 45.12)(200, 45.25)(400, 45.25)(800, 47.33)
            };
        \addlegendentry{BYOL~\cite{grill2020byol}}
                \addplot[
            color=color_blue,
            mark=x,
            line width=0.4mm,
            ]
            coordinates {
            (100, 45.15)(200, 45.08)(400, 46.16)(800, 47.62)
            };
        \addlegendentry{DenseCL~\cite{Wang2021DenseCL}}
        \addplot[
            color=color_orange,
            mark=diamond,
            line width=0.4mm,
            ]
            coordinates {
            (100, 43.62)(200, 44.36)(400, 45.38)(800, 46.51)
            };
        \addlegendentry{InsLoc~\cite{Yang2021InstanceLF}}
        \addplot[
            color=color_pink,
            mark=x,
            line width=0.4mm,
            ]
            coordinates {
            (100, 44.48)(200, 45.67)(400, 45.74)(800, 46.04)
            };
        \addlegendentry{Agri BT~\cite{roggiolani2023icra-odsp}}
    \end{axis}
    \end{tikzpicture}}

\caption{Ablation study on the impact of pretraining for an increasing number of epochs. The results are averaged over three runs.}
\label{fig:ablation-epochs}
\end{figure}
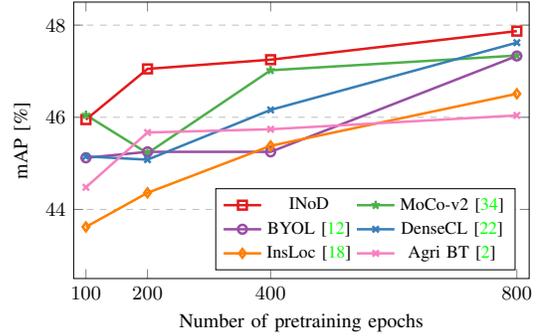
In this section, we study the impact of pretraining schedule length on the overall performance of the model. We pretrain all the SSL baselines for $100$, $200$, $400$, and $800$ epochs and report the corresponding finetuning results in~\figref{fig:ablation-epochs}. We observe that \net~consistently outperforms the SSL baselines when pretrained with longer training schedules and also approaches its final convergence score faster than the other baselines. For instance, our model achieves an mAP of $47.05$\% at $200$ epochs which only marginally improves to $47.87$\% when pretrained for $800$ epochs. In contrast, the baselines achieve an improvement of \rebuttal{\SI{1.85}{pp}} on average when the training schedule is increased four-fold from $200$ epochs to $800$ epochs. We emphasize that quick and early convergence is a crucial attribute for SSL-based approaches as it reduces the use of computational resources and improves practicality for many real-world applications. 

\subsubsection{Granularity of Noise Injection}
\label{ablation:granularity}

\begin{table}
\centering
\caption{Ablation study on the granularity of noise injection. All metrics are reported in [$\%$] and averaged over three runs.}
\label{tab:ablation-ref}
\footnotesize
\setlength\tabcolsep{3.7pt}
 \begin{tabular}{c|cccc}
 \toprule
 \textbf{Granularity} & \textbf{1/4} & \textbf{1/8} & \textbf{1/16} & \textbf{1/32}\\
 \midrule
 mAP & \textbf{47.05} & 46.71 & 45.00 & 46.58 \\
 \bottomrule
 \end{tabular}
 \vspace{-0.3cm}
\end{table}

\begin{table}[t]
\centering
\caption{Ablation study on varying the amount of noise injection. All metrics are reported in [$\%$] and averaged over three runs}
\label{tab:ablation-nr}
\footnotesize
\setlength\tabcolsep{3.7pt}
 \begin{tabular}{c|cc|c}
 \toprule
 \textbf{Amount of Injection} & \textbf{AP\textsubscript{crop}} & \textbf{AP\textsubscript{weed}} & \textbf{mAP} \\
 \midrule
$10\%$ & 59.33 & 32.55 & 45.94 \\
$20\%$ & \textbf{60.85} & 33.24 & \textbf{47.05} \\
$30\%$ & 58.58 & 33.02 & 45.80 \\
$40\%$ & 59.76 & \textbf{33.46} & 46.61 \\
 \bottomrule
 \end{tabular}
\vspace{-0.3cm}
\end{table}

We analyze the impact of the granularity of noise injection during pretraining in~\tabref{tab:ablation-ref}. As defined in \secref{subsec:noise-mask-generation}, granularity defines the smallest possible volume that can be replaced in the source feature map. Using the smallest granularity of $\frac{1}{4}$ (smallest replaceable unit of $4$ pixels) results in the highest performance. Consequently, we argue that injecting noise with a high resolution (low granularity) allows the network to learn precise features pertaining to the edges of plant structures, thus improving the object detection performance of the model.

\begin{figure*}
\centering
\begin{subfigure}[b]{0.495\textwidth}
\vskip 0pt
\centering
\footnotesize
\setlength{\tabcolsep}{0.05cm}
{
\renewcommand{\arraystretch}{0.2}
\newcolumntype{M}[1]{>{\centering\arraybackslash}m{#1}}
\begin{tabular}{cM{1.95cm}M{1.95cm}M{1.95cm}M{1.95cm}}
& Input image & DenseCL~\cite{Wang2021DenseCL} & \net~(Ours) & Improv./Error \\
(a) & \includegraphics[width=\linewidth, frame]{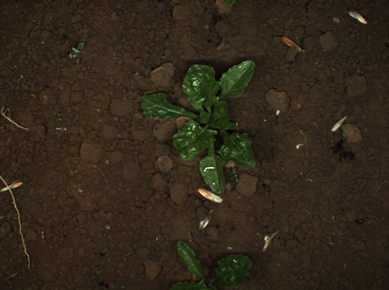} & \includegraphics[width=\linewidth, frame]{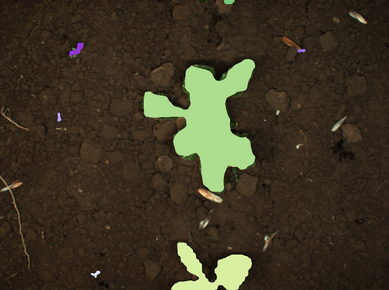} & \includegraphics[width=\linewidth, frame]{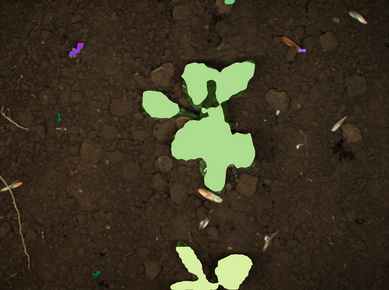} & \includegraphics[width=\linewidth, frame]{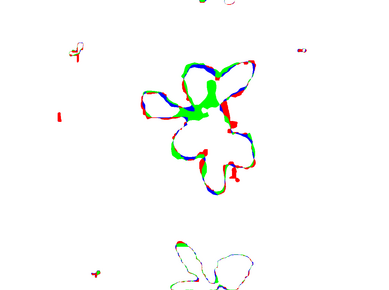}\\
\\
(b) &  \includegraphics[width=\linewidth, frame]{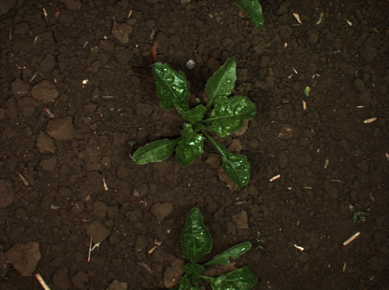} & \includegraphics[width=\linewidth, frame]{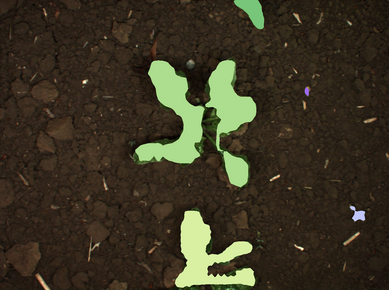} & \includegraphics[width=\linewidth, frame]{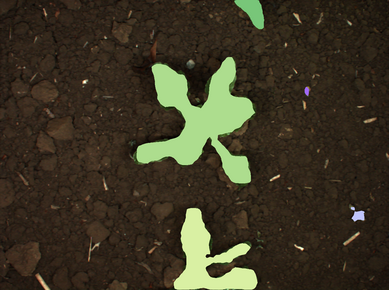} & \includegraphics[width=\linewidth, frame]{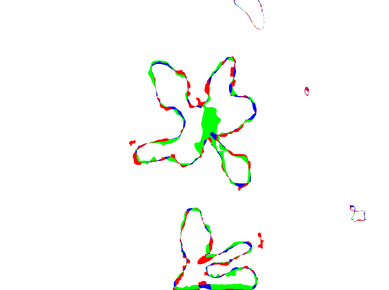}\\
\\
(c) & \includegraphics[width=\linewidth, frame]{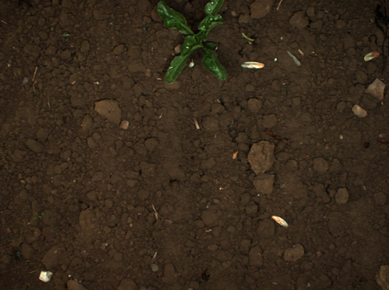} & \includegraphics[width=\linewidth, frame]{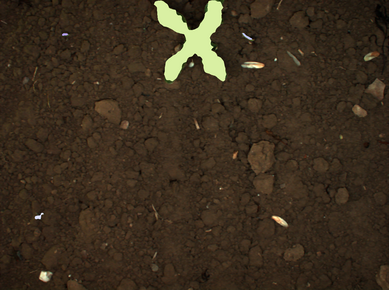} & \includegraphics[width=\linewidth, frame]{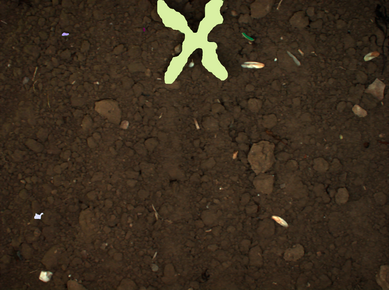} & \includegraphics[width=\linewidth, frame]{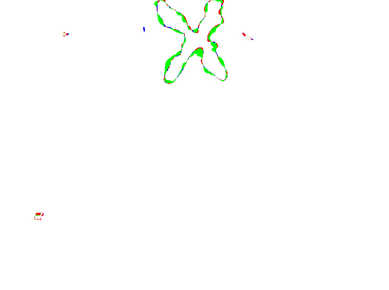} \\
\\
(d) & \includegraphics[width=\linewidth,  frame]{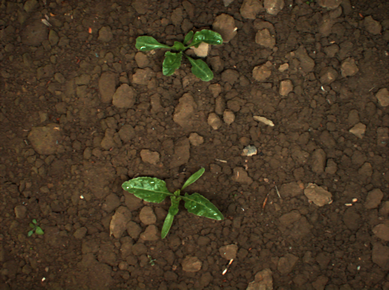} & \includegraphics[width=\linewidth, frame]{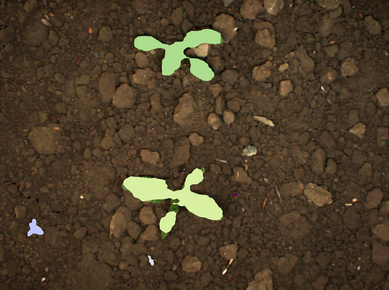} & \includegraphics[width=\linewidth, frame]{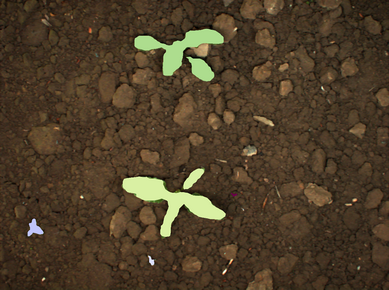} & \includegraphics[width=\linewidth, frame]{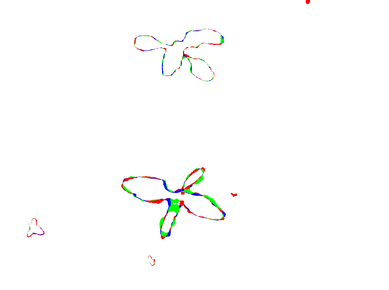} \\
\end{tabular}
}
\caption{Instance segmentation results. Crop and weed instances are colored using varying shades of green and purple respectively.}
\label{fig:qual-analysisIS}
\end{subfigure}
\hfill
\begin{subfigure}[b]{0.495\textwidth}
\vskip 0pt
\centering
\footnotesize
\setlength{\tabcolsep}{0.05cm}
{
\renewcommand{\arraystretch}{0.2}
\newcolumntype{M}[1]{>{\centering\arraybackslash}m{#1}}
\begin{tabular}{cM{2cm}M{2cm}M{2cm}M{2cm}}
& Input image & MoCo-v2~\cite{chen2020mocov2} & \net~(Ours) & Improv./Error \\
(a) & \includegraphics[width=\linewidth, frame]{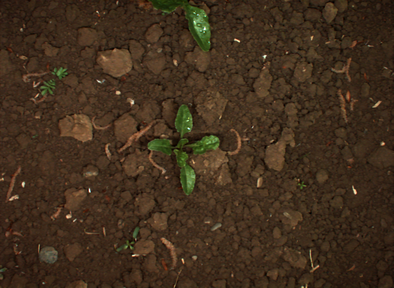} & \includegraphics[width=\linewidth, frame]{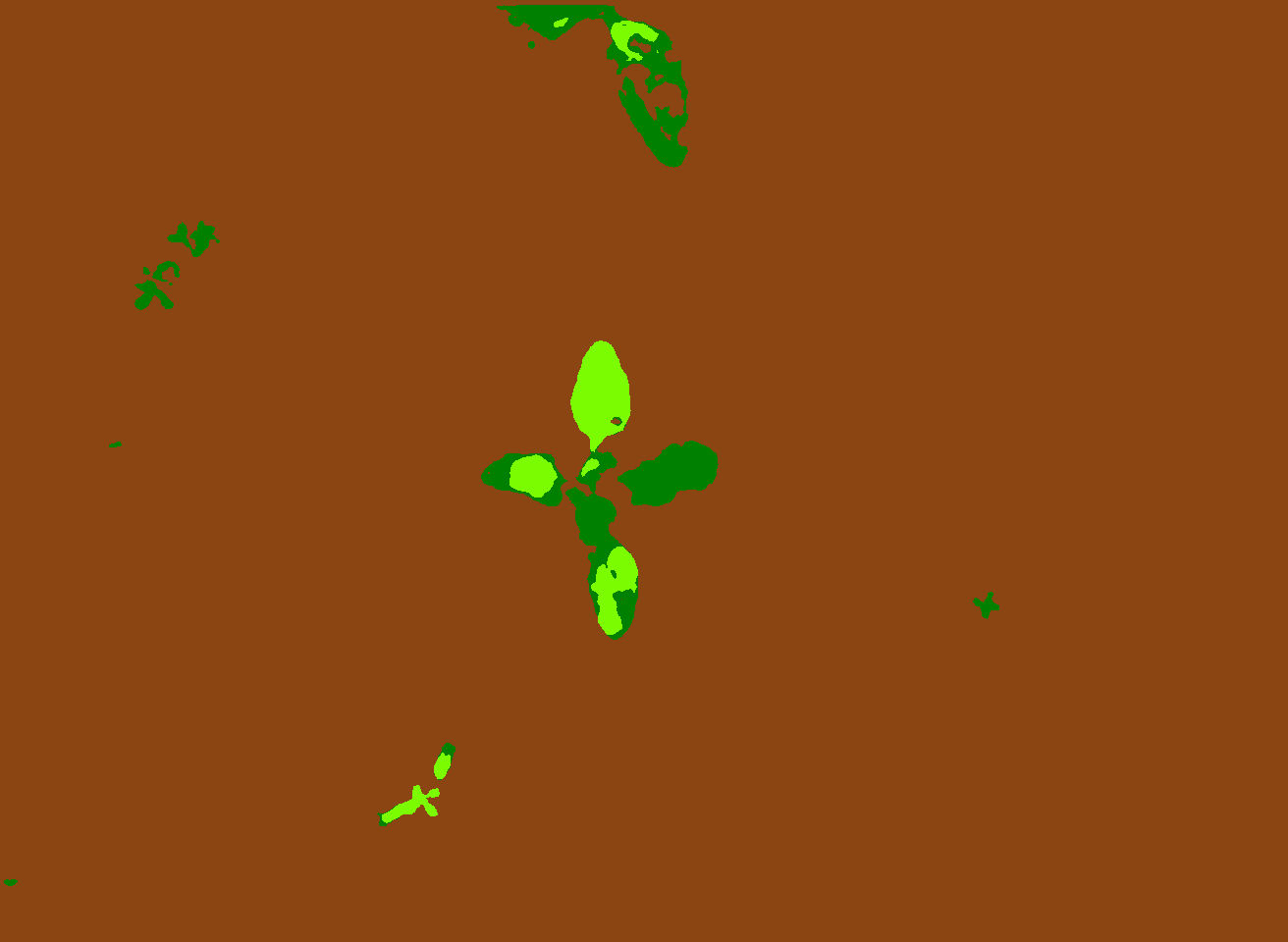} & \includegraphics[width=\linewidth, frame]{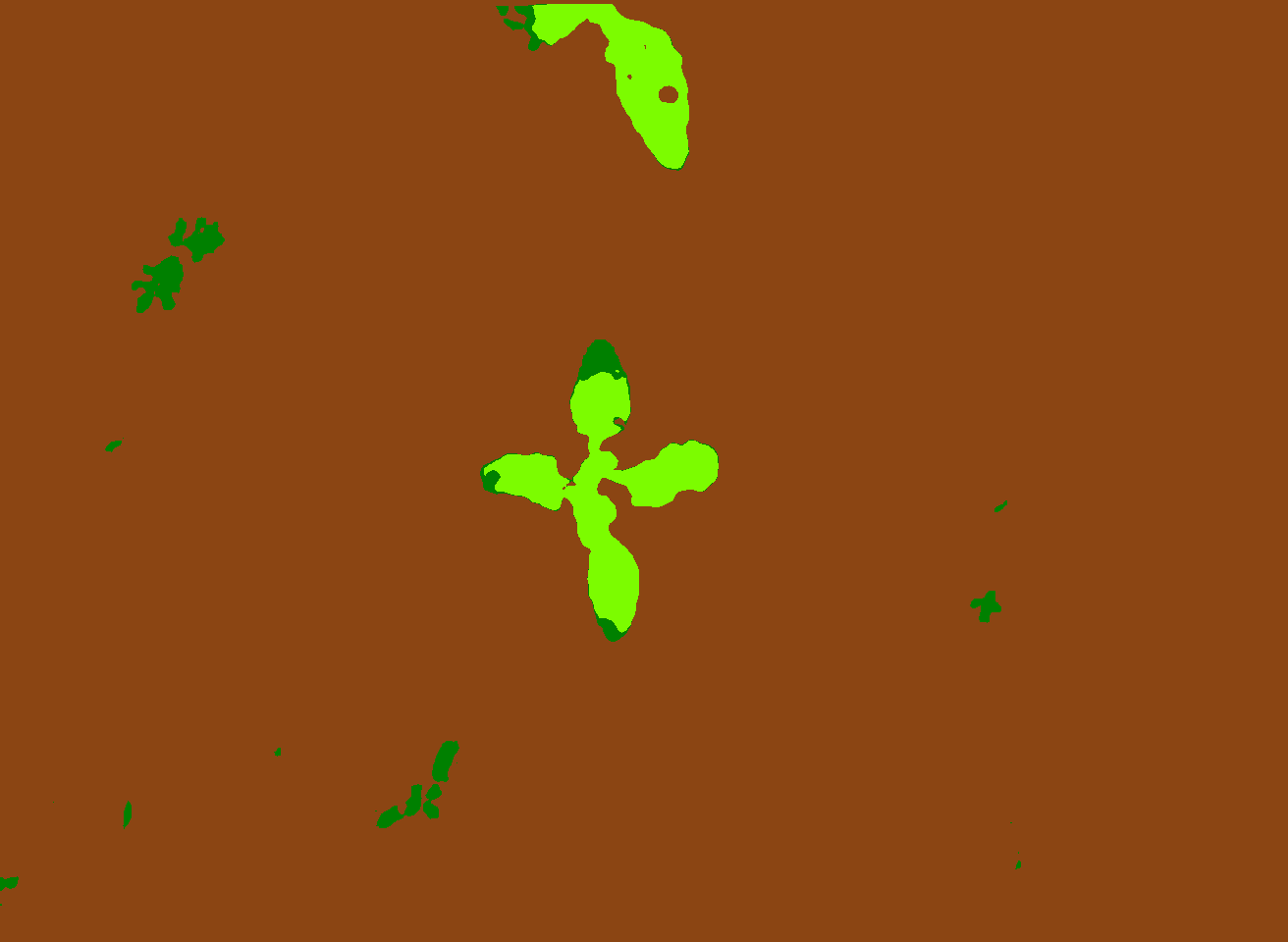} & \includegraphics[width=\linewidth, frame]{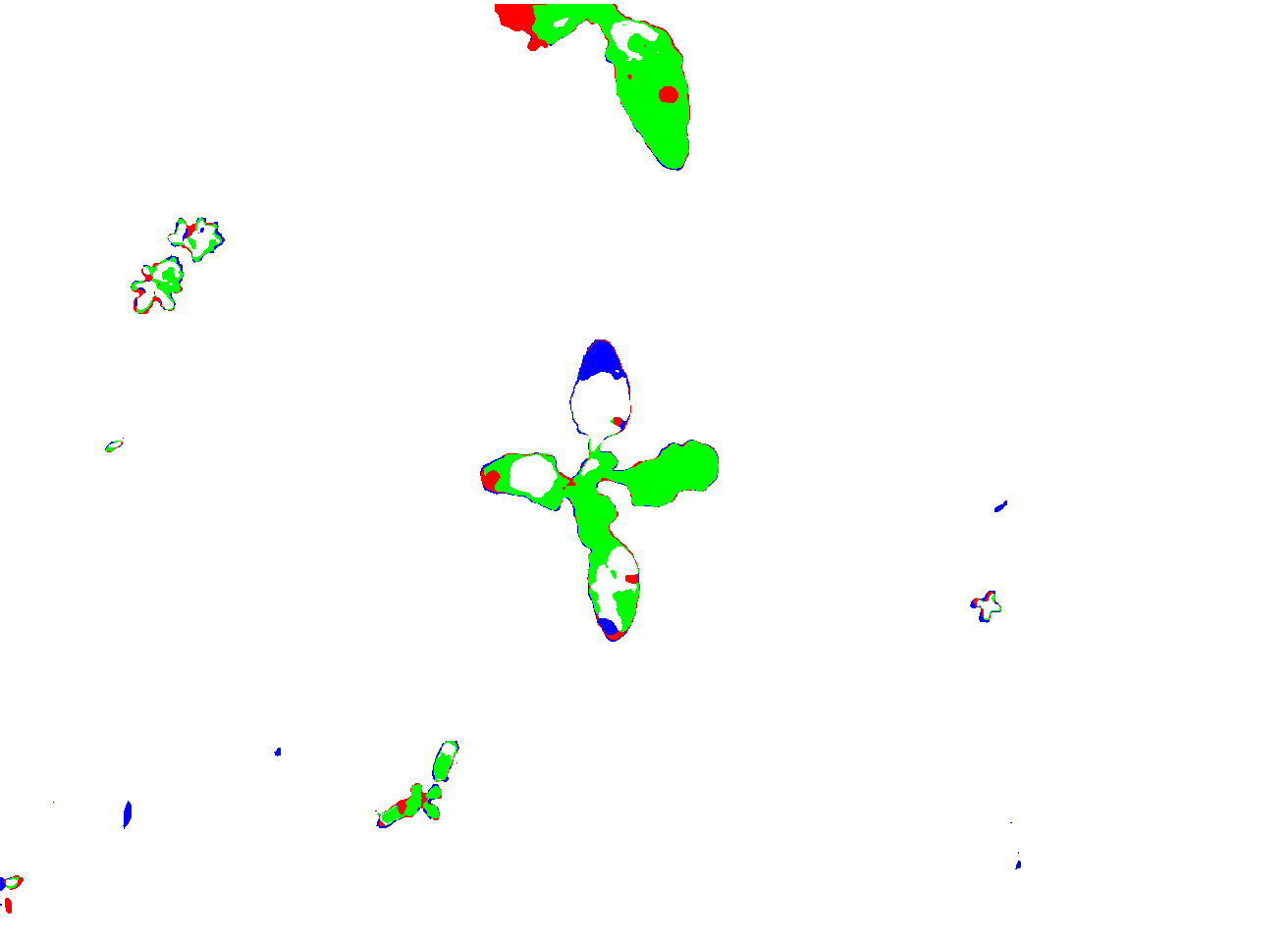}\\
\\
(b) &  \includegraphics[width=\linewidth, frame]{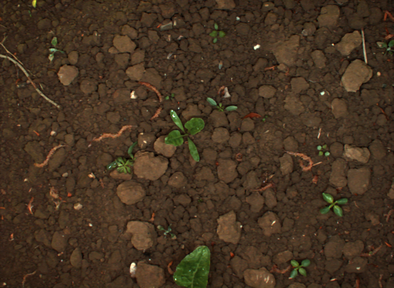} & \includegraphics[width=\linewidth, frame]{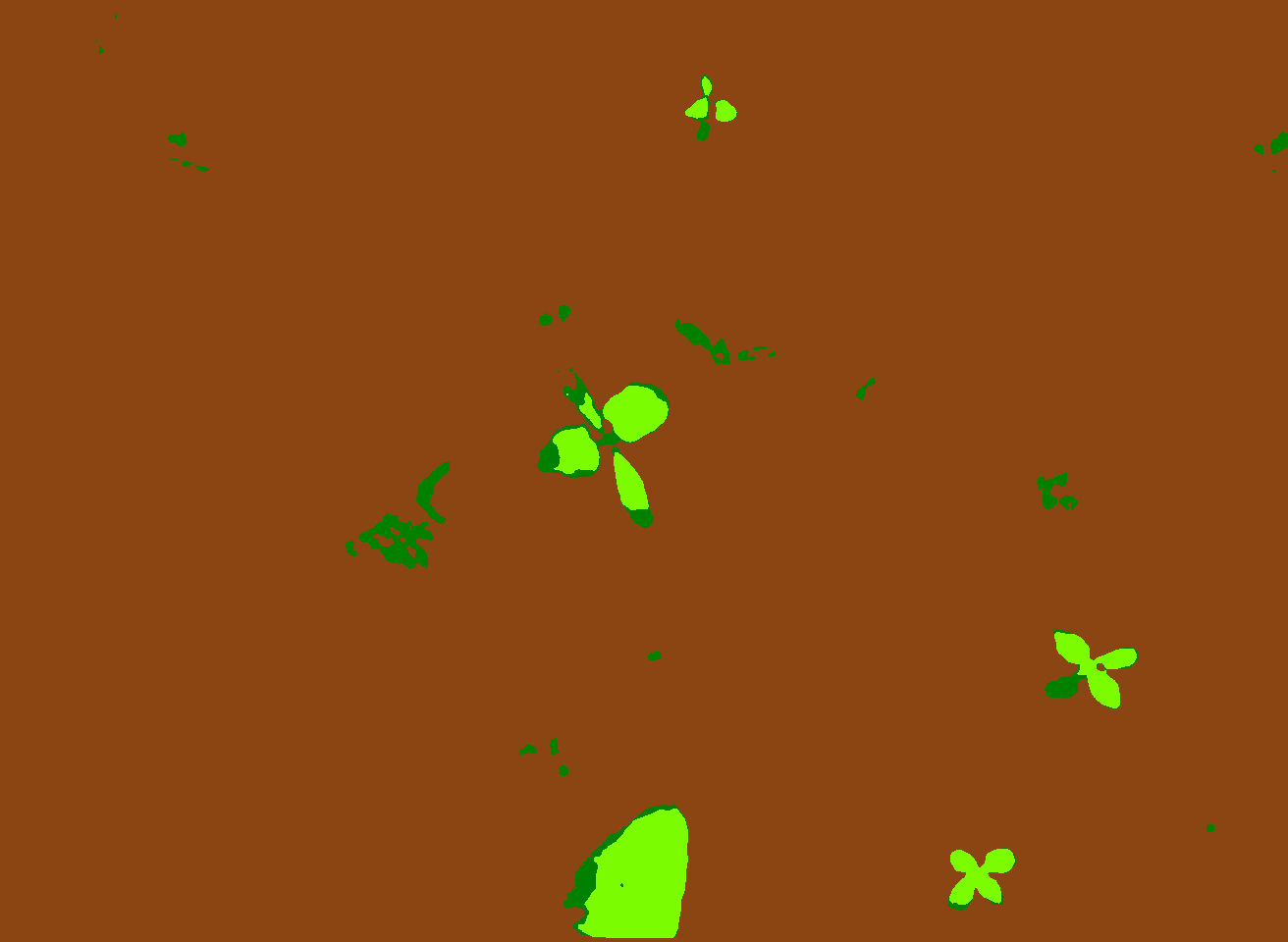} & \includegraphics[width=\linewidth, frame]{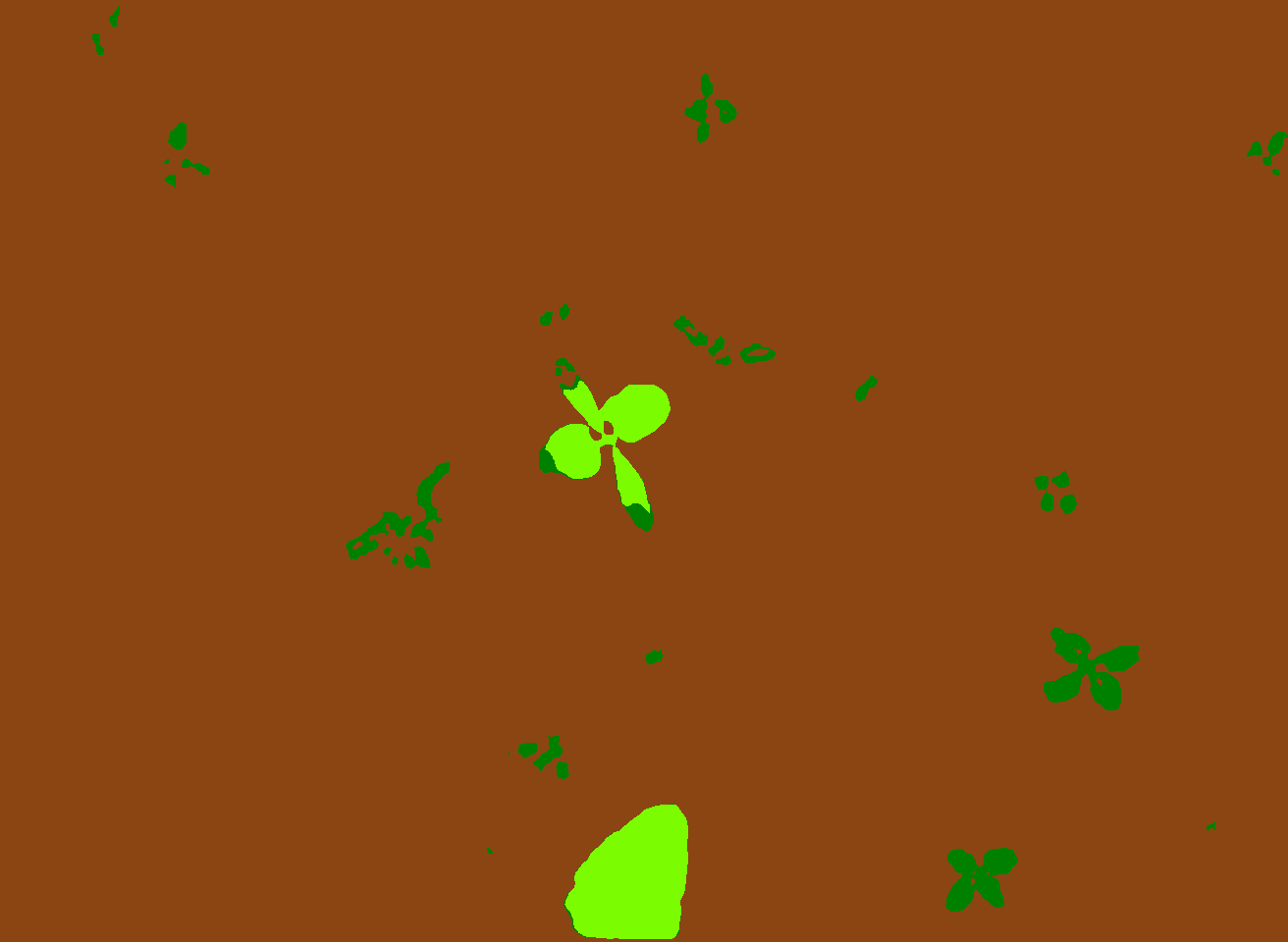} & \includegraphics[width=\linewidth, frame]{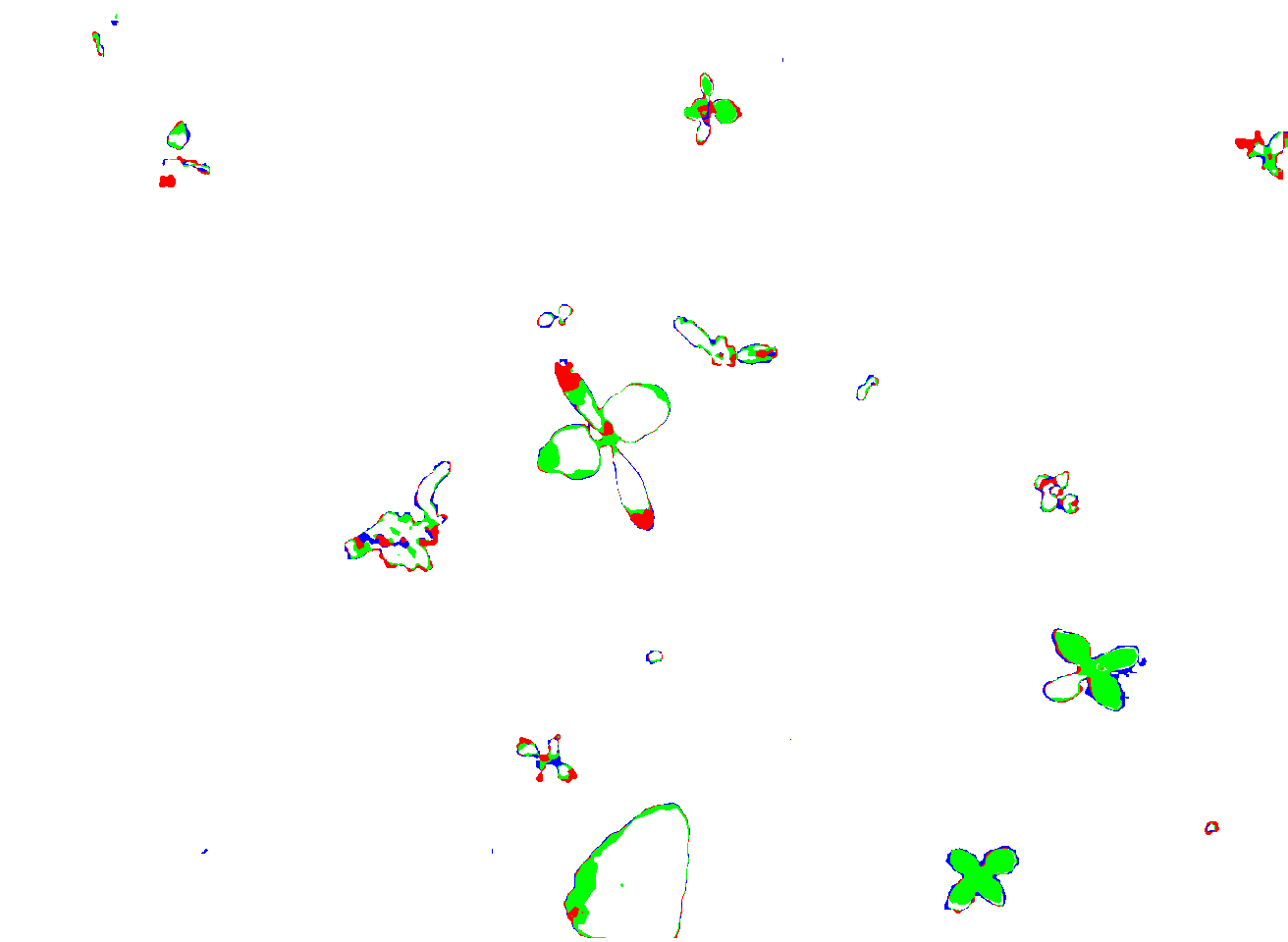}\\
\\
(c) & \includegraphics[width=\linewidth, frame]{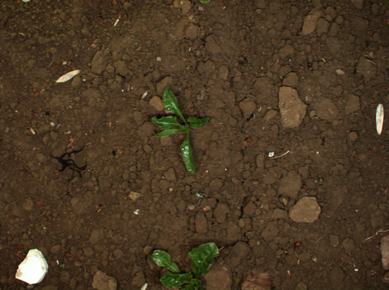} & \includegraphics[width=\linewidth, frame]{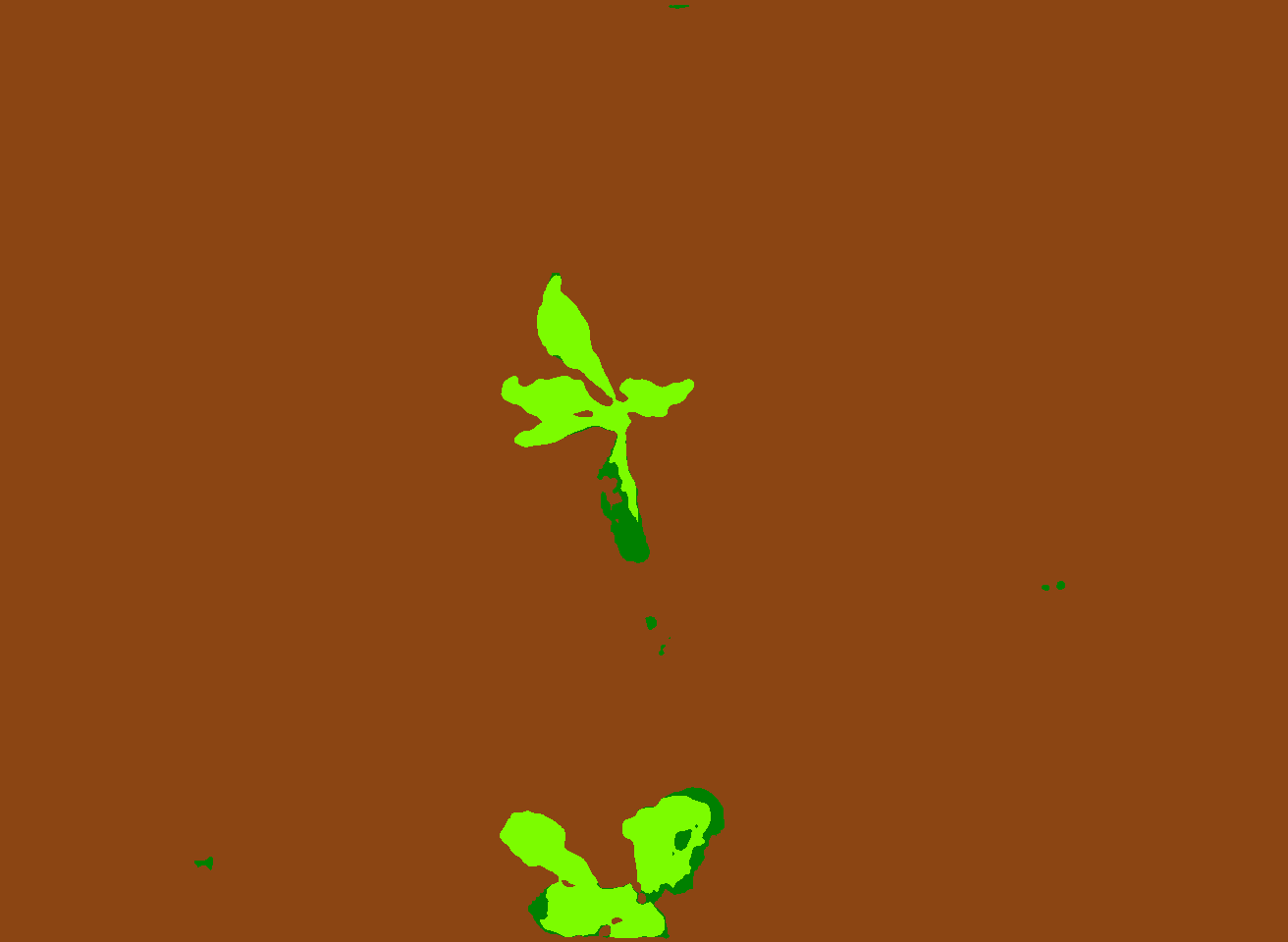} & \includegraphics[width=\linewidth, frame]{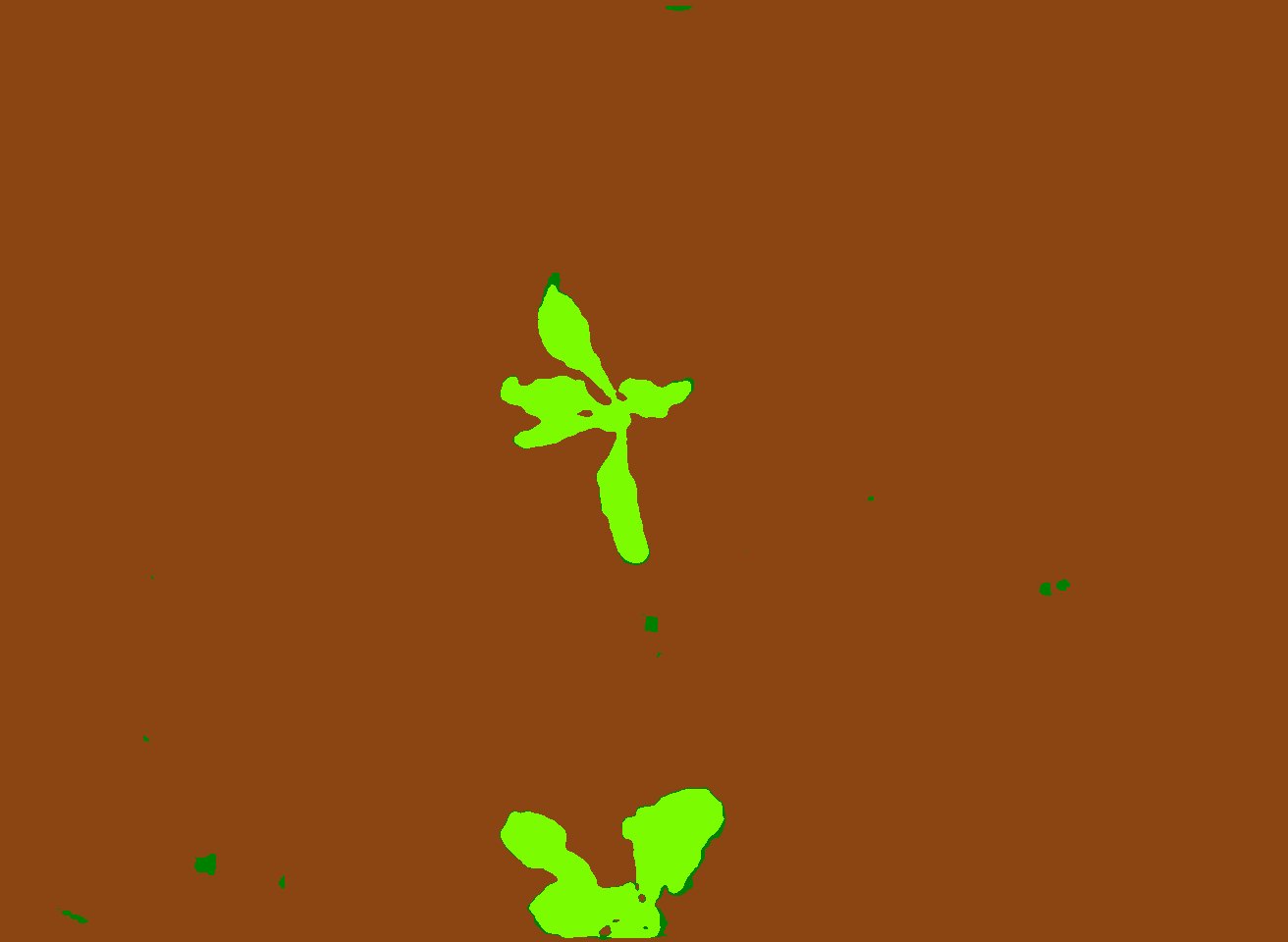} & \includegraphics[width=\linewidth, frame]{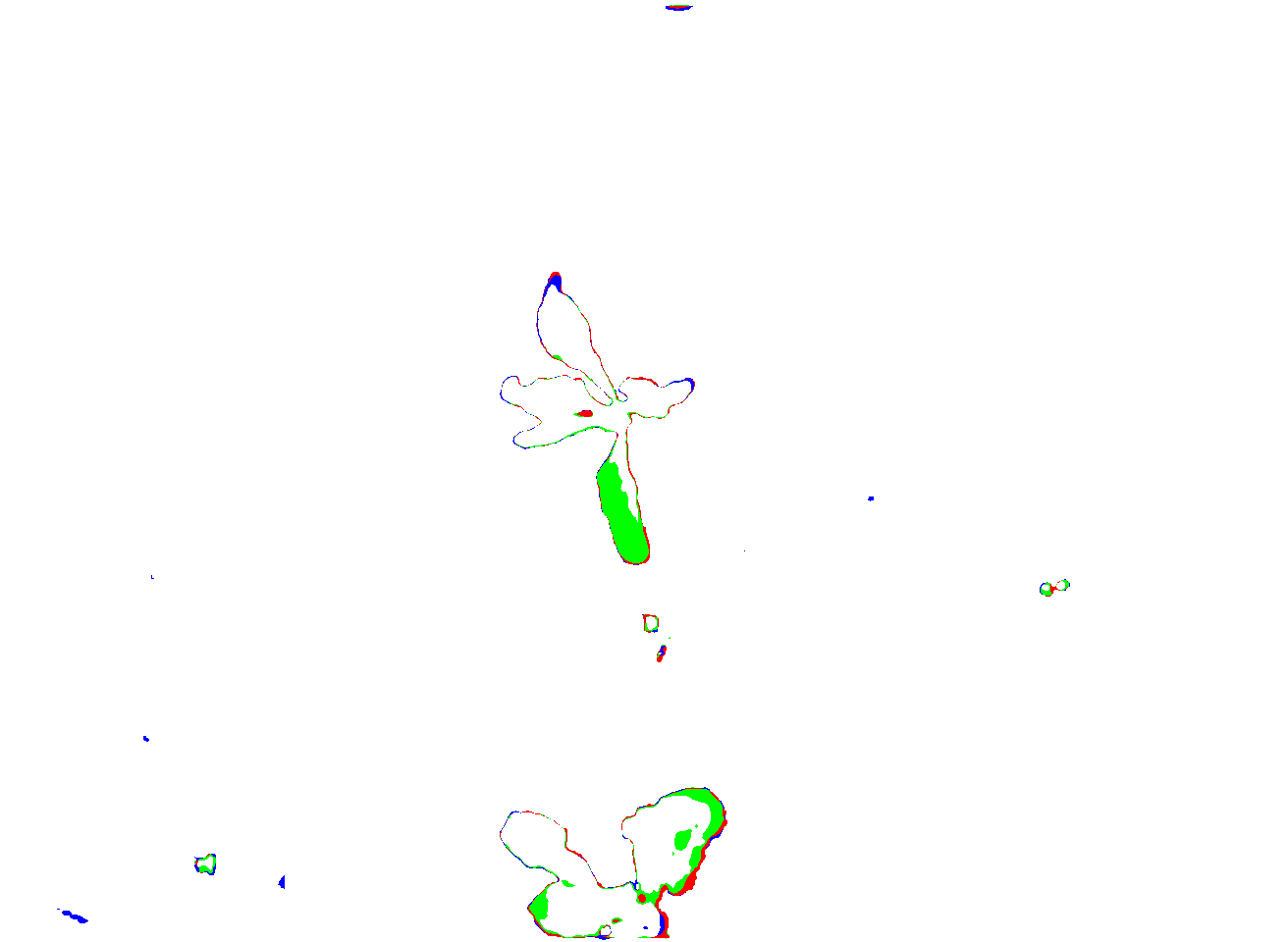} \\
\\
(d) & \includegraphics[width=\linewidth,  frame]{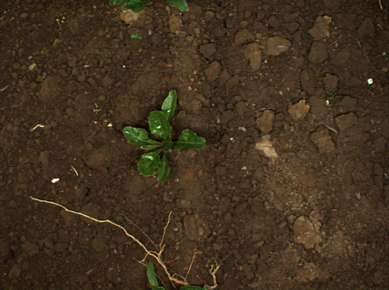} & \includegraphics[width=\linewidth, frame]{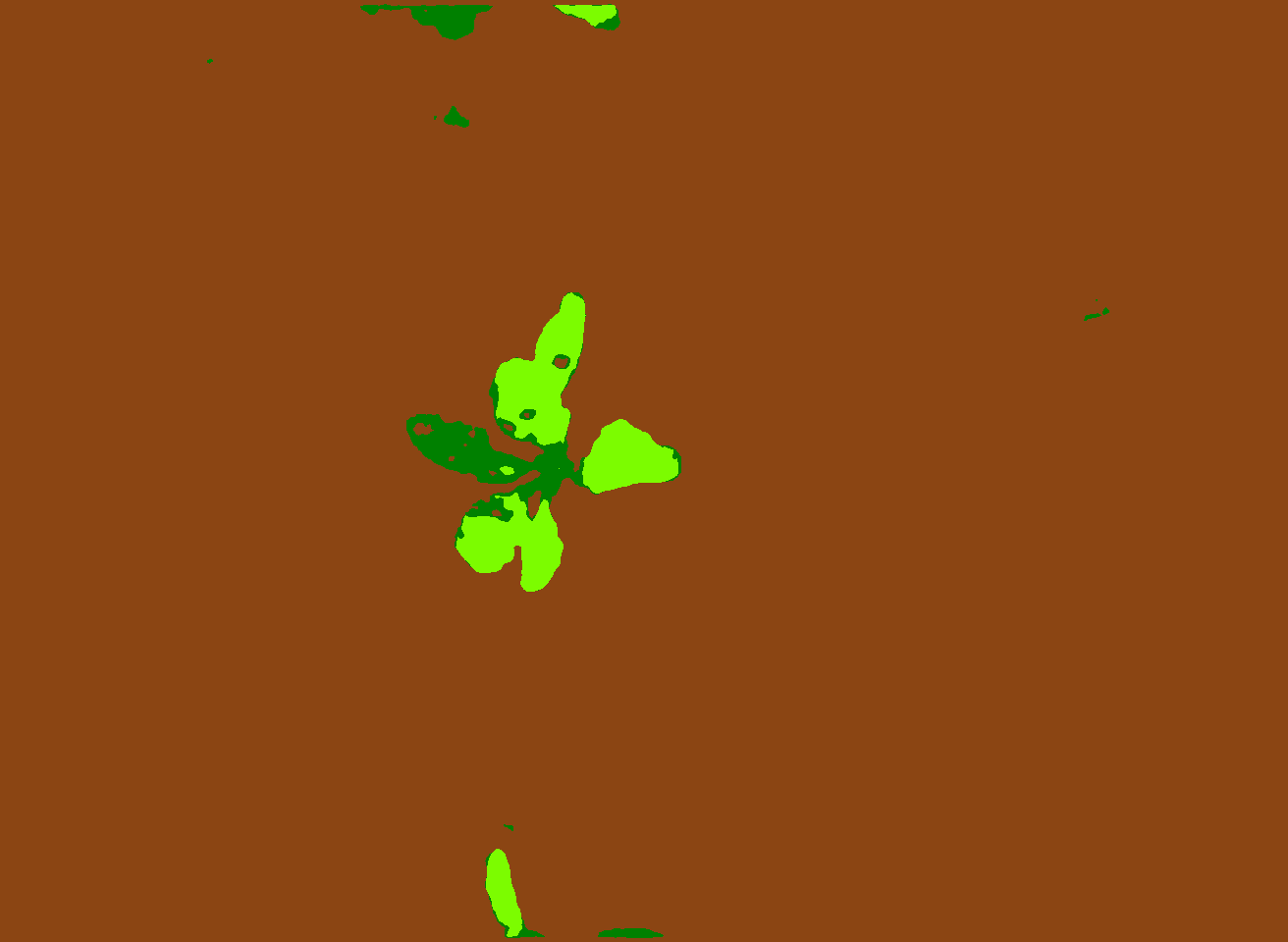} & \includegraphics[width=\linewidth, frame]{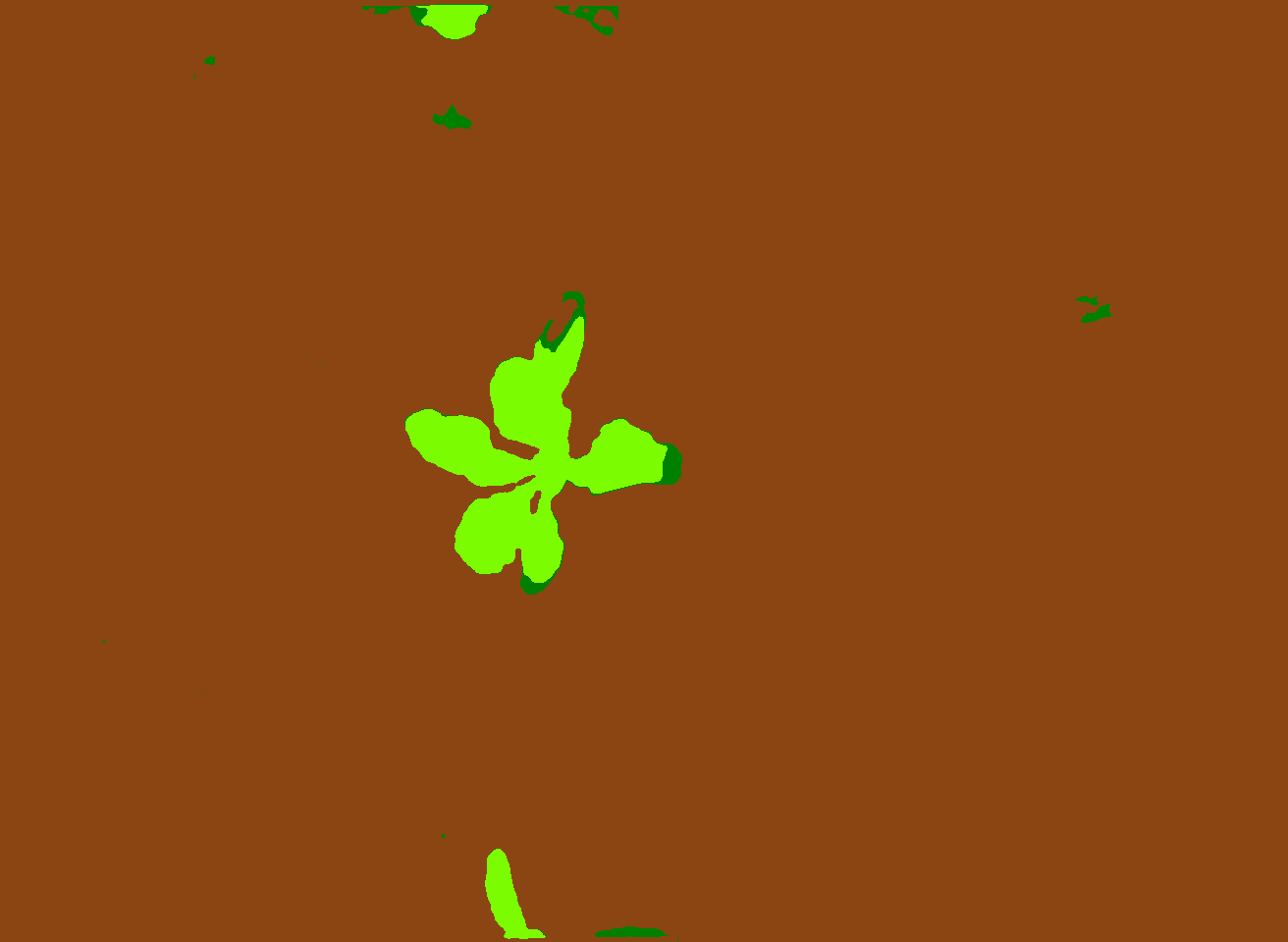} & \includegraphics[width=\linewidth, frame]{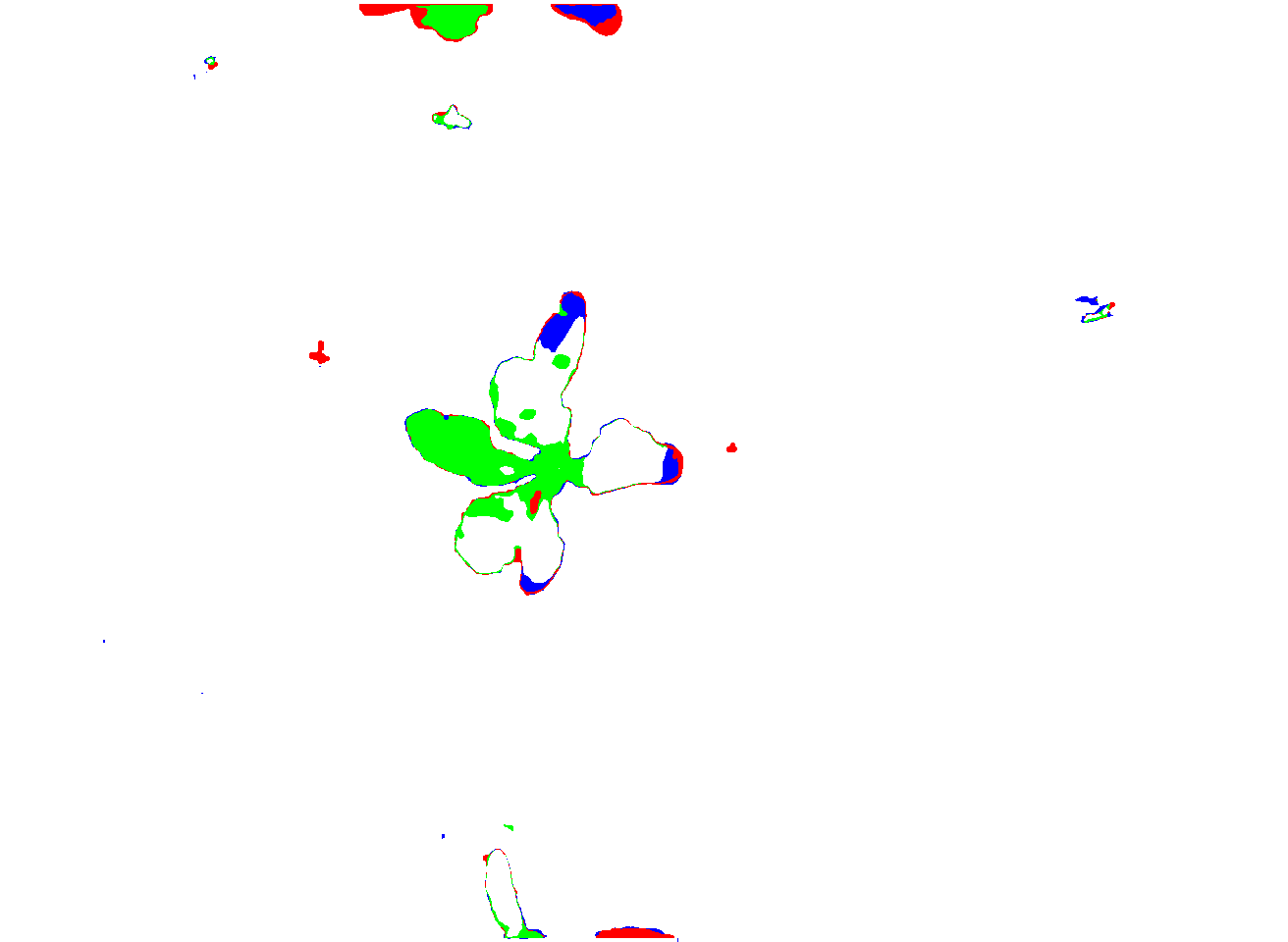} \\
\end{tabular}
}
\caption{Semantic segmentation results. Crops are shown in light green, weeds in dark green, and soil in brown.}
\label{fig:qual-analysisSS}
\end{subfigure}
\caption{Qualitative results of \net~in comparison to the best performing baselines for instance and semantic segmentation on the SB16 dataset. The improvement/error maps show pixels misclassified by the baseline and correctly predicted by \net~in green and vice-versa in blue. Incorrect predictions of both models are colored in red.}
\end{figure*}

\begin{figure*}
\centering
\footnotesize
\setlength{\tabcolsep}{0.05cm}
{
\renewcommand{\arraystretch}{0.2}
\newcolumntype{M}[1]{>{\centering\arraybackslash}m{#1}}
\begin{tabular}{M{2.6cm}M{2.6cm}M{2.6cm}M{2.6cm}M{2.6cm}M{2.6cm}}
\includegraphics[width=\linewidth, frame]{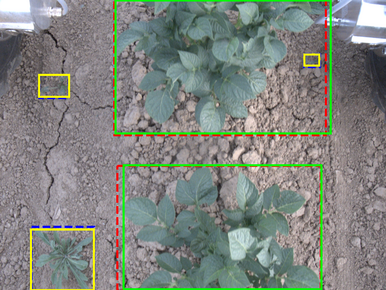} & 
\includegraphics[width=\linewidth, frame]{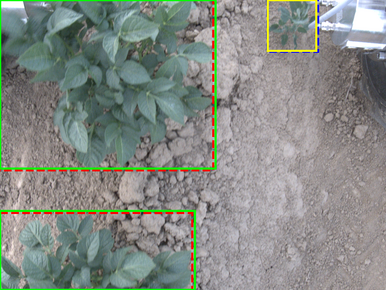} & 
\includegraphics[width=\linewidth, frame]{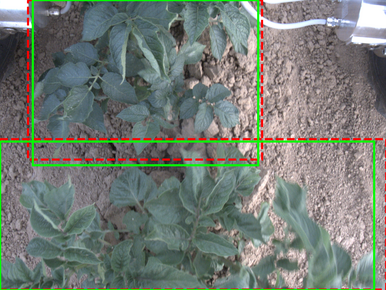} &
\includegraphics[width=\linewidth, frame]{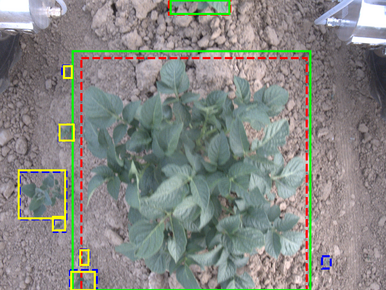} & \includegraphics[width=\linewidth, frame]{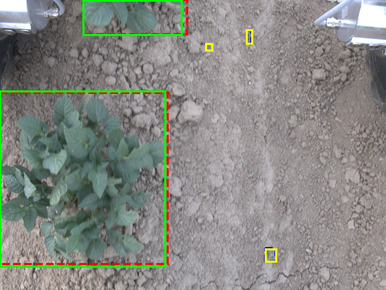}
& \includegraphics[width=\linewidth, frame]{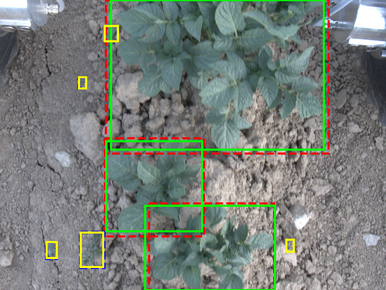} \\ \\ 
(a)& (b) & (c) & (d) & (e) & (f)\\
\end{tabular}
}
\caption{Qualitative results of our self-supervised \net~framework on our Fraunhofer Potato 2022 dataset.}
\label{fig:qual-cp}
\vspace{-0.3cm}
\end{figure*}

\subsubsection{Quantity of Noise Injection}
\label{ablation:quantity}

In this section, we study the impact of varying amounts of noise injection during the pretraining step of our \net~approach. Accordingly, we replace $10$\%, $20$\%, $30$\%, and $40$\% \rebuttal{of} source features with noise features and present the \rebuttal{results on the target task} in  \tabref{tab:ablation-nr}. We observe that the model performance slightly deteriorates when we replace either $10$\%, $30$\% or $40$\% of source features with noise features compared to the model where we \rebuttal{replace} $20$\% of source features with noise. A small amount of noise injection makes the pretraining task very easy and hinders the network from learning rich and meaningful features. Consequently, the finetuning step is also hindered which results in slightly worse results for both crop and weed classes. In contrast, a large amount of noise injection results in fewer observed crops belonging to the source dataset during pretraining, which in turn dampens the learning of the model for the crop class. However, weeds exhibit similar characteristics across the source and noise datasets which minimizes the impact of high noise injection for the weed class. Nevertheless, we highlight that the amount of noise injection is a key parameter which should be carefully tuned according to the task and dataset.

\subsection{Qualitative Results} 
\label{qual}
We further evaluate the impact of \net~by qualitatively comparing its predictions with the best performing SSL baseline for the instance and semantic segmentation tasks in \figref{fig:qual-analysisIS} and \figref{fig:qual-analysisSS}, respectively. For instance segmentation, we observe that both DenseCL and \net~enable the network to segment all multi-leaf crops in the image while effectively neglecting dead vegetation in the image extremities. In \figref{fig:qual-analysisIS}(c), we note that \net~achieves a finer segmentation of the potato crop compared to DenseCL which is evident from the improvement/error map shown in the last column. Further, we observe in \figref{fig:qual-analysisIS}(a,\rebuttal{~b,}~d) that \net~is able to segment intricate crop structures and also the connections between individual leaves, while DenseCL fails to do so. Moreover, we also highlight that \net~is able to better segment crops at the boundaries of images as shown in \figref{fig:qual-analysisIS}(b). We argue that this characteristic stems from our pretraining approach that discourages the mixing of convolutional features in the backbone which results in the precise localization of features, especially at image borders where padding is applied. Additionally, we note that both pretraining approaches struggle to identify small weeds which are shown as red dots in the error maps of \figref{fig:qual-analysisIS}(a,~b,\rebuttal{~d}).

In the semantic segmentation results shown in \figref{fig:qual-analysisSS}, we observe that both models precisely segment vegetation from soil but often misclassify crop and weed plants. While \net~routinely predicts the edges of crop leaves as weed, MoCo-v2 commonly misclassifies entire leaves as shown in \figref{fig:qual-analysisSS}(a, c, d). This characteristic can be attributed to our data split during finetuning wherein the training set comprises images from early farming days having smaller plant structures while the test set contains mature plants. Consequently, MoCo-v2 \rebuttal{fails} to adapt to this change in plant sizes while \net~\rebuttal{succeeds}, thus highlighting the generalizability of our approach. Lastly, we note that MoCo-v2 commonly identifies small weeds as crops in \figref{fig:qual-analysisSS}(a, b) while \net~better generalizes on such occurrences.

We also qualitatively evaluate the object detection performance on the FP22 dataset in \figref{fig:qual-cp}. Here, the crop and weed predictions are colored in green and yellow, while the ground truth labels are colored in red and blue, respectively. We observe that \net~predicts tight bounding boxes around complete plant structures and correctly classifies the detected objects. However, we observe that our model occasionally fails to accurately determine the exact edges of overlapping plants as shown in \figref{fig:qual-cp}(c, d). Similar to the instance segmentation results, our model also fails to detect small weeds as is evident from the blue bounding box having no corresponding yellow box in \figref{fig:qual-cp}(d). Nevertheless, as demonstrated in \figref{fig:qual-cp} our model detects most crops and weeds reliably thus enabling its use in a multitude of precision agriculture applications.
\section{Conclusion}
\label{sec:conclusion}

In this paper, we present \net, a novel self-supervised pretraining strategy for dense prediction tasks in agricultural domains. Our approach is based on incorporating principles of feature replacement and dataset discrimination during the convolutional encoding of two disjoint datasets.
We observe that our pretraining strategy outperforms the existing state-of-the-art SSL baselines as well as ImageNet pretraining for object detection, semantic segmentation, and instance segmentation on the SB16 dataset as well as our new FP22 dataset. We also publicly release our novel FP22 dataset comprising $16,800$ images for object detection in potato fields. Our approach is one of the early works in SSL to provide a targeted pretraining solution for multiple dense prediction tasks, without architecture-specific customization. Consequently, this research motivates the development of novel SSL approaches outside of the highly explored but limited contrastive learning objective.


                                  

{\footnotesize
\bibliographystyle{IEEEtran}
\bibliography{references}
}
\clearpage

\begin{strip}
\begin{center}
\vspace{-5ex}
\textbf{\Large \bf
INoD: Injected Noise Discriminator for Self-Supervised Representation Learning in Agricultural Fields
} \\
\vspace{2ex}

\Large{\bf- Supplementary Material -}\\
\vspace{0.4cm}
\normalsize{Julia Hindel, Nikhil Gosala, Kevin Bregler and Abhinav Valada}
\end{center}
\end{strip}

\setcounter{section}{0}
\setcounter{equation}{0}
\setcounter{figure}{0}
\setcounter{table}{0}
\makeatletter

\renewcommand{\thesection}{S.\arabic{section}}
\renewcommand{\thesubsection}{S.\arabic{section}.\Alph{subsection}}
\renewcommand{\thetable}{S.\arabic{table}}
\renewcommand{\thefigure}{S.\arabic{figure}}

\normalsize
In this supplementary material, we present additional illustrations of our methodology in \secref{sec:vis} and further experiments to study the performance of \net~in \secref{sec:quan}. Lastly, we also outline additional implementation details in \secref{sec:impl}.

\section{Additional Illustrations of Noise Mask and Pseudo Label Generation} \label{sec:vis}

\begin{figure*}
\vskip 0pt
\centering
\footnotesize
\setlength{\tabcolsep}{0.05cm}
{
\renewcommand{\arraystretch}{0.2}
\newcolumntype{M}[1]{>{\centering\arraybackslash}m{#1}}
\begin{tabular}{M{2.6cm}M{2.6cm}M{2.6cm}M{2.6cm}M{2.6cm}}
Reference Noise Mask ($N$) & $N_1$(res2) & $N_2$(res3) & $N_3$(res4) & $N_4$(res5)\\
\includegraphics[width=\linewidth, frame]{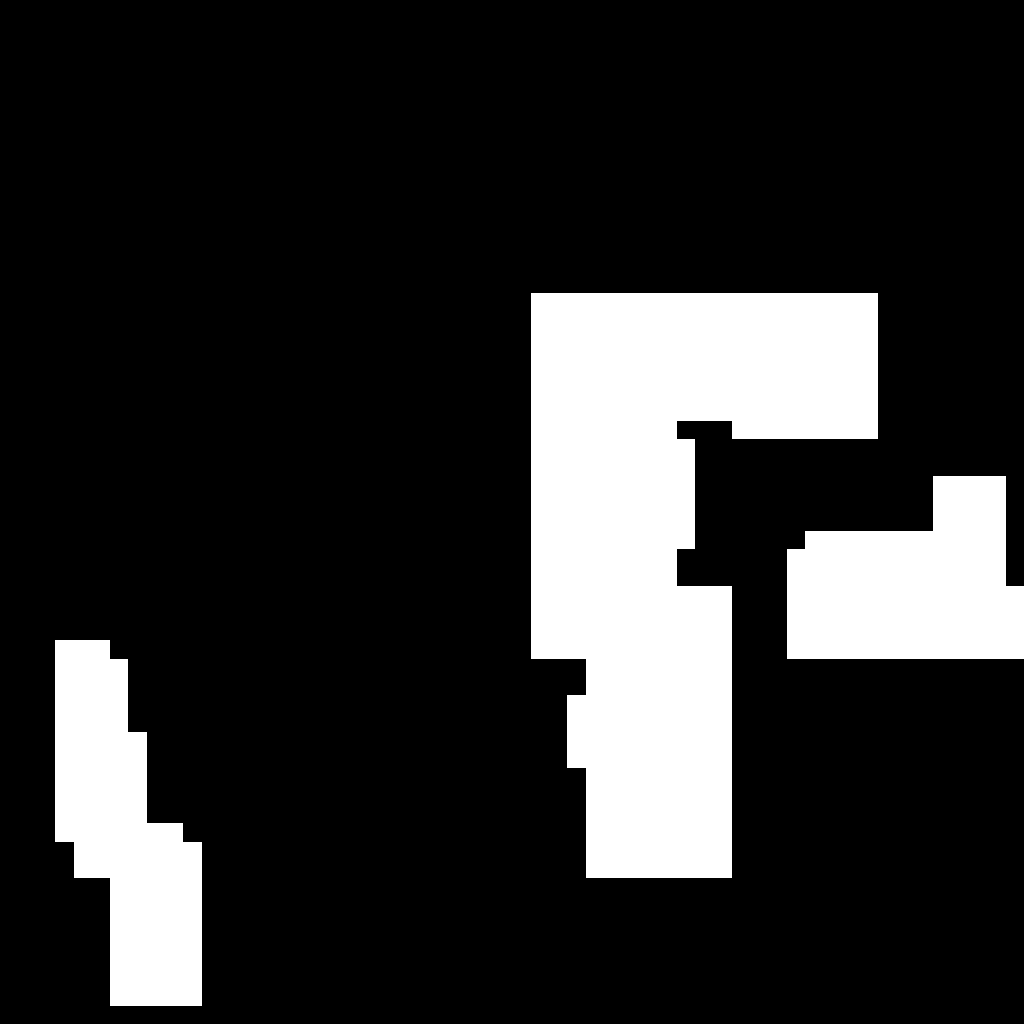} & 
\includegraphics[width=\linewidth, frame]{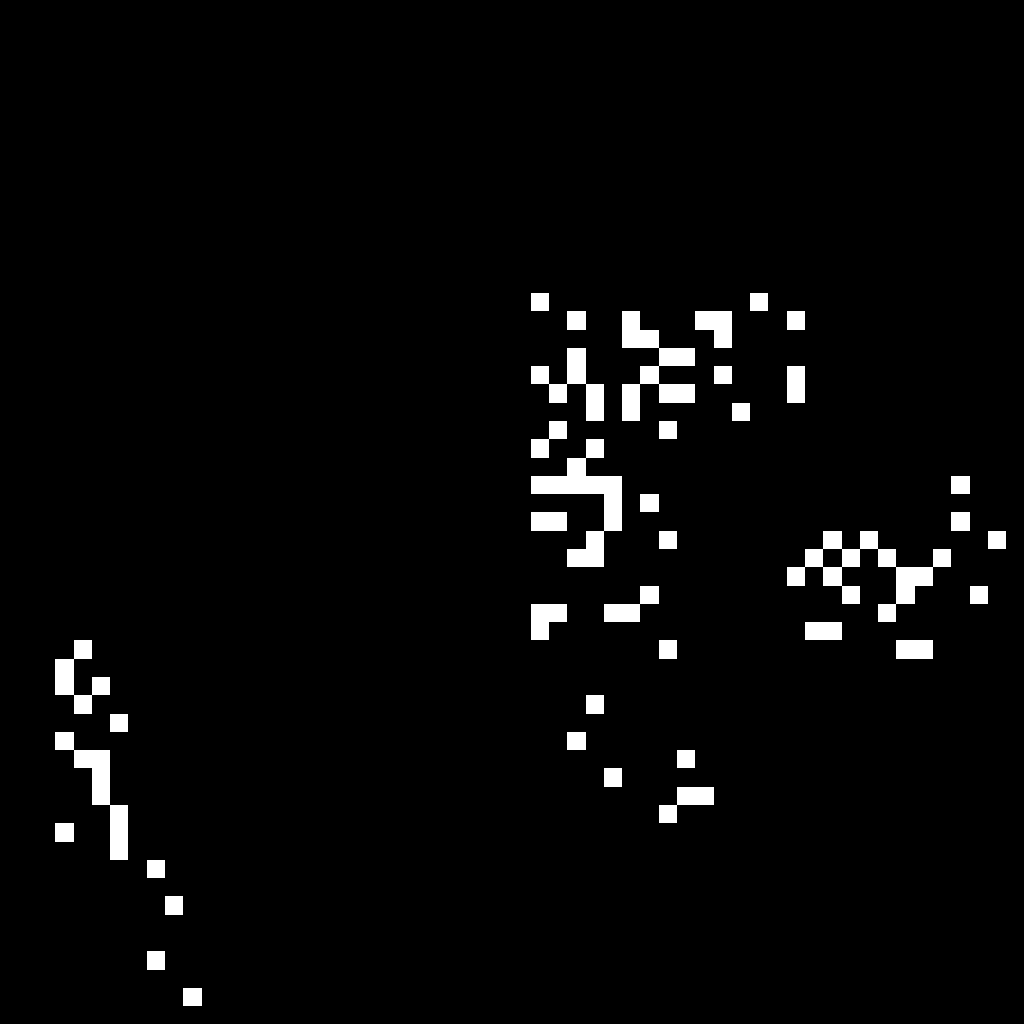} & 
\includegraphics[width=\linewidth, frame]{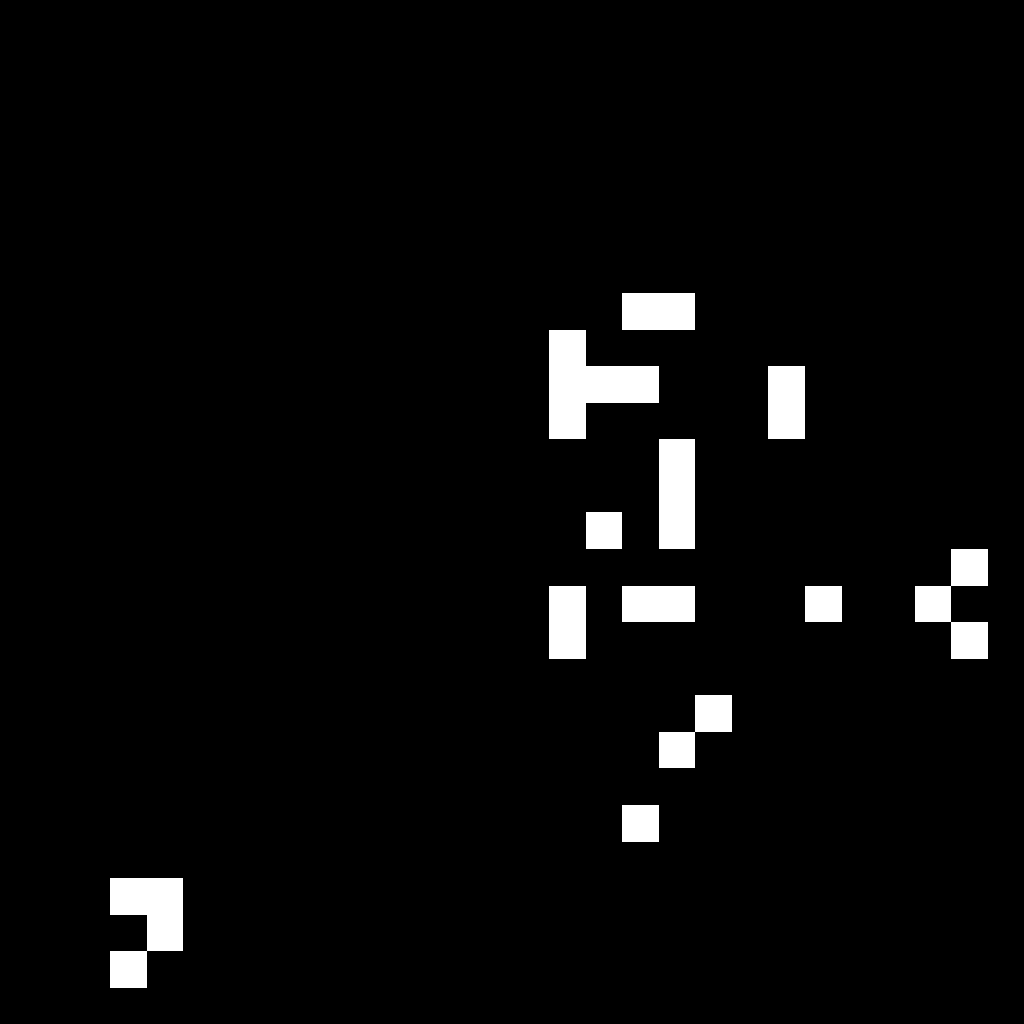} & 
\includegraphics[width=\linewidth, frame]{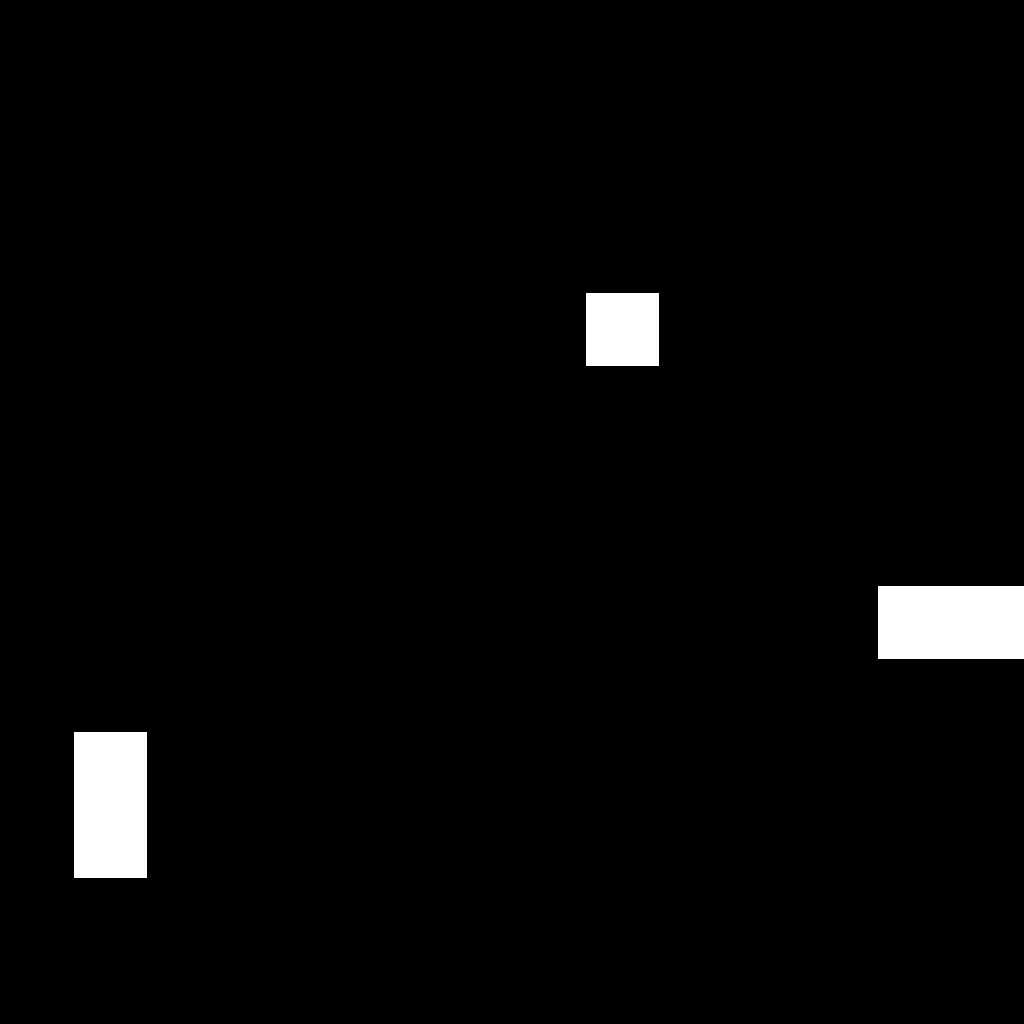} & 
\includegraphics[width=\linewidth, frame]{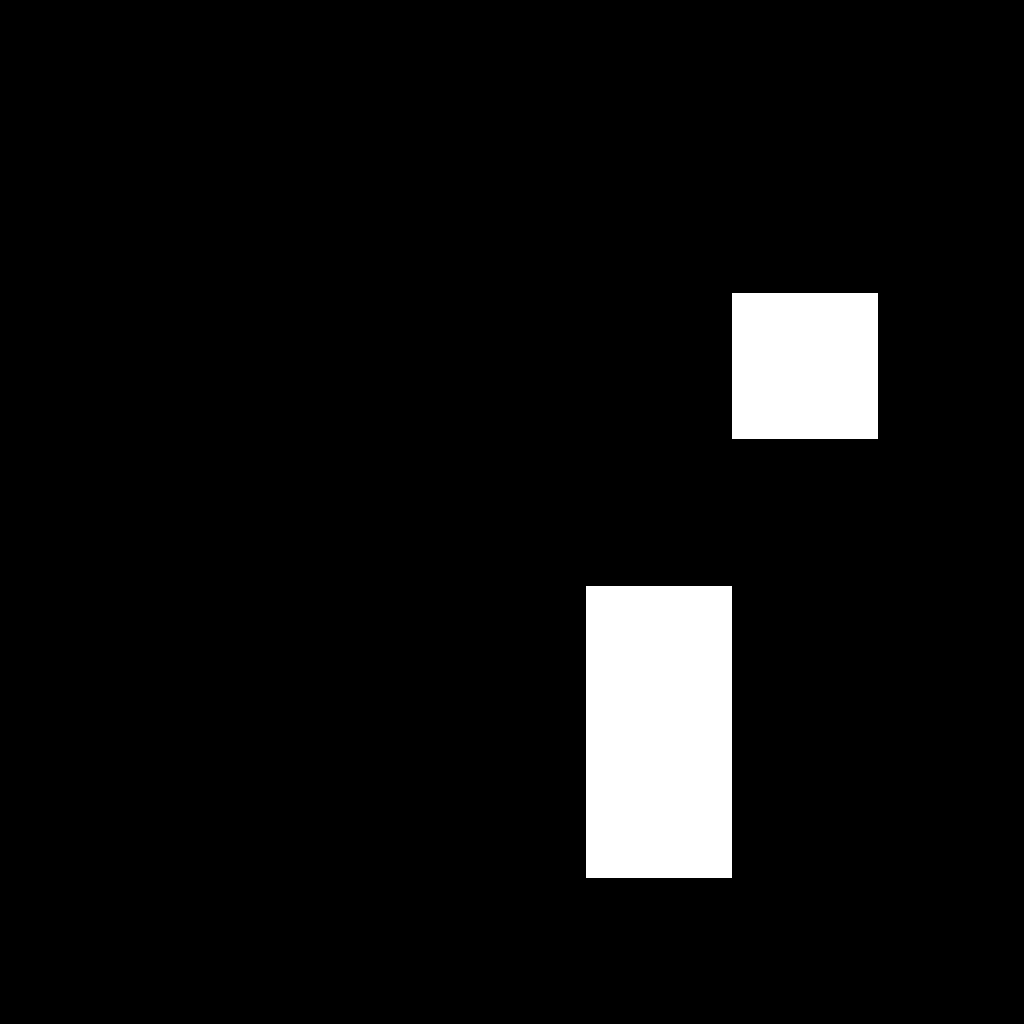}\\
\\
\end{tabular}
}
\caption{Visualization of reference noise mask ($N$) split into layer-specific masks ($N_{1..4}$).}
\label{fig:noise-mask}
\end{figure*}

\figref{fig:noise-mask} illustrates the random division of a reference noise mask, $N$, into four layer-specific noise masks, $N_{1...4}$, as detailed in \secref{subsec:noise-injection}. In \figref{fig:noise-mask}, $N_{1}$ corresponds to noise injection after the res2 block of a ResNet-50 backbone, where we replace feature chunks having a granularity of $\frac{1}{4}$  (smallest replaceable unit of $4$ pixels). In the later stages of the convolutional backbone, the granularity of the noise injection regions is coarser which results in the replacement of larger effective views. For example, the smallest replaceable sizes for noise masks $N_{2}$, $N_{3}$, and $N_{4}$ are $\frac{1}{8}$, $\frac{1}{16}$, and $\frac{1}{32}$ respectively. In \net, we always inject noise after every residual block in the ResNet-50 backbone and demonstrate the benefit of including feature injection with a granularity of $\frac{1}{4}$ in \secref{ablation:granularity}.

\begin{figure*}
\centering
   	\includegraphics[width=0.75\linewidth]{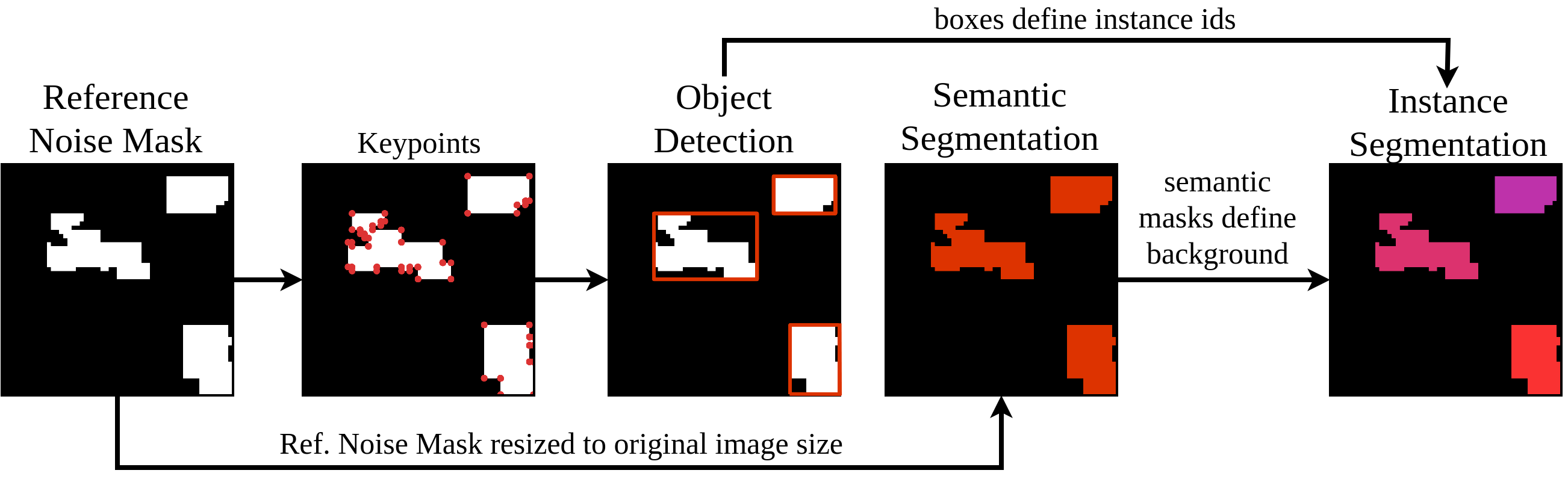}
\caption{Visualization of pseudo label creation for each task.}
\label{fig:labels}
\end{figure*}

Further, we display the pseudo label computation for all three tasks in \figref{fig:labels}. To retrieve object detection pseudo labels, we extract bounding boxes around individual contours of noise areas. We compute semantic segmentation labels by resizing the reference noise mask to the original image size. Lastly, we generate the instance segmentation labels by combining the masked pixels from the semantic segmentation pseudo labels and instance IDs from object detection pseudo labels.

\section{Additional Quantitative Results} \label{sec:quan}
We analyze the performance of \net~in comparison to its benchmarks on the target task of object detection on SB16 in \secref{sec:suppdetec}, and observe the impact of using ImageNet-initialized backbones for pretraining in \secref{sec:suppin}. Further, we perform additional ablative experiments to study the characteristics of our contribution in relation to different noise datasets in \secref{sec:suppnoise}, and dissimilar pretraining and target tasks in \secref{sec:ssl-transfer}.

\subsection{Object detection on SB16} \label{sec:suppdetec} 
We present the object detection scores from finetuning the Mask R-CNN architecture on Sugar Beets 2016 in \tabref{tab:sb16rcnn}. We employ equivalent pretraining and finetuning settings as described in \secref{subsec:experimental-settings}. Our proposed model, \net, outperforms its baselines by at least \SI{3.23}{pp} on this task. While the boosted performance can be attributed to superior detection scores on both crop and weed classes,  it is worth noting that the AP$_{crop}$ score of \net~exceeds the best-performing baseline by \SI{6.32}{pp}. The consistent enhancements observed in instance segmentation, semantic segmentation, and object detection for this dataset support our claim that \net~significantly improves crop recognition through its semantically meaningful pretraining.

\begin{table}
\footnotesize
\centering
\caption{Evaluation of object detection on Sugar Beets 2016. All metrics are reported in [$\%$] and averaged over three runs.}
\label{tab:sb16rcnn}
\setlength\tabcolsep{3.7pt}
 \begin{tabular}{l|cc|ccc}
 \toprule
 \textbf{Pretraining} & \textbf{AP\textsubscript{crop}} & \textbf{AP\textsubscript{weed}} &\textbf{mAP} & \textbf{AP\textsubscript{75}} & \textbf{AP\textsubscript{50}} \\
 \midrule
Supervised~(IN) & 57.09 & 18.84 & 37.96 & 42.54 & 53.44  \\
MoCo-v2~\citeS{S-chen2020mocov2} & 62.29 & 21.81 & 42.05 & 47.76 & 57.63 \\
BYOL~\citeS{S-grill2020byol} & 59.44 & 20.09 & 39.77 & 45.38 & 56.49 \\ 
InsLoc~\citeS{S-Yang2021InstanceLF} & 57.75 & 18.43 & 38.09 & 43.04 & 52.16 \\
DenseCL~\citeS{S-Wang2021DenseCL} & 63.14 & 22.11 & 42.63 & 48.99 & 60.00 \\ 
AgriBT~\citeS{S-roggiolani2023icra-odsp} & 56.44 & 19.54 & 37.99 & 43.45 & 52.83 \\
\cmidrule{1-6}
\net~(Ours) & \textbf{69.46} & \textbf{22.25} & \textbf{45.86} & \textbf{50.95} & \textbf{61.42} \\
\bottomrule
\end{tabular}
\vspace{-0.2cm}
\end{table}

\subsection{ImageNet 
Initialization} \label{sec:suppin}
In \tabref{tab:in-init}, we evaluate the performance achieved on object detection with FP22 when the ResNet-50 backbone is initialized with ImageNet weights prior to self-supervised pretraining. 
We observe that initializing the pretraining models using ImageNet weights as opposed to random initialization results in an increased performance on the target task w.r.t. the mAP metric for all the tested baselines. This increase in performance can be primarily attributed to improved detection on the weed class - which is also a consequence of using ImageNet, as evident from the ImageNet row in \tabref{tab:fastrcnn}. We also note that all self-supervised pretraining approaches complement the ImageNet weight initialization and achieve better results as compared to their random initialization counterparts. Further, our \net~approach consistently outperforms all the baselines in the mAP metric even when initialized using ImageNet pretrained weights which demonstrates the robustness of our approach to different pretraining settings.

\begin{table}
\footnotesize
\centering
\caption{Evaluation of object detection on the Fraunhofer Potato 2022 dataset when the pretraining model uses ImageNet initialization and Random initialization. All metrics are reported in [$\%$] and averaged over three runs.}
\label{tab:in-init}
\setlength\tabcolsep{3.7pt}
 \begin{tabular}{l|ccc|ccc}
 \toprule
 & \multicolumn{3}{c|}{\textbf{ImageNet Intialization}} & \multicolumn{3}{c}{\textbf{Random Intialization}}\\
 \textbf{Pretraining} & \textbf{AP\textsubscript{crop}} & \textbf{AP\textsubscript{weed}} &\textbf{mAP} & \textbf{AP\textsubscript{crop}} & \textbf{AP\textsubscript{weed}} &\textbf{mAP}\\
 \midrule
MoCo-v2~\citeS{S-chen2020mocov2} & 59.24 & 35.87 & 47.56 & 58.05 & 32.39 & 45.22\\
BYOL~\citeS{S-grill2020byol} & 57.69 & \textbf{36.75} & 47.22 & 56.46 & \textbf{34.04} & 45.25 \\
InsLoc~\citeS{S-Yang2021InstanceLF} & 57.91 & 36.54 & 47.23 & 58.31 & 30.40 & 44.36 \\
DenseCL~\citeS{S-Wang2021DenseCL} & 58.83	& 35.80 & 47.32 & 57.63 & 32.54 & 45.08 \\ 
AgriBT~\citeS{S-roggiolani2023icra-odsp} & 56.43 & 36.44 & 46.44 & 57.70 & 33.64 & 45.67 \\
\cmidrule{1-7}
\net~(Ours) & \textbf{61.76} & 36.63 & \textbf{49.19}& \textbf{60.85} & 33.24 & \textbf{47.05}\\
\bottomrule
\end{tabular}
\vspace{-0.2cm}
\end{table}

\subsection{Noise dataset} \label{sec:suppnoise}
We evaluate the impact on target performance when applying different datasets as noise. For this ablation, we use SB16 as a source dataset and employ held-out samples from SB16, Synthetic SB16~\citeS{fppbnRAS2021}, FP22, OPDD~\citeS{pheno2020}, MinneApple~\citeS{hani2019minneapple}, and Cityscapes~\citeS{Cordts2016TheCD} as the noise dataset where each noise dataset becomes progressively dissimilar to the source dataset as compared to its predecessor. We display sample images of each dataset in \figref{fig:samples}. Further, we also evaluate the performance of the model when the injected noise are random pixels having the same distribution (mean and standard deviation) as the source dataset. We present the target performance on object detection and instance segmentation on SB16 in \tabref{tab:noisedataset}.

We observe from \tabref{tab:noisedataset} that the object detection metrics exhibit higher dependency on the choice of noise dataset as compared to the instance segmentation task as evident from the high variance in the mAP score. We also note that the AP\textsubscript{crop} score in object detection is highly influenced by this pretraining parameter, while the AP\textsubscript{weed} metric does not exhibit a high variance. 
For both tasks, using a held-out split of the source dataset as noise results in the lowest downstream performance ($\downarrow$~\SI{6.83}{pp} mAP object detection, $\downarrow$~\SI{2.18}{pp} mAP instance segmentation). We highlight that the network fails to differentiate between features if they are from the same origin which results in the network not learning meaningful features, and subsequently in poor performance on the target task.
In comparison, utilizing similar agricultural datasets (synthetic dataset produced by conditional GAN network (Synthetic SB16) or another tuber plant (FP22)) results in superior performance on both object detection and instance segmentation. We conclude that this setting enables the network to focus on fine differences between related plants to resolve the pretraining task which positively affects the performance on the target task.
Further, using an out-of-domain dataset such as Cityscapes can lead to improved performance on the crop class in comparison to the weakly-related MinneApple dataset and OPPD. We argue that Cityscapes aids in refining feature encodings due to its comparably high diversity.
Lastly, we note that employing random pixels with the same dataset distribution (mean and standard deviation) as the source dataset provides a promising alternative if no suitable noise dataset is available. We reason that artificial noise enables the learning of semantically meaningful features as it possesses greater diversity. This configuration also outperforms our baselines on the mAP metric for object detection and instance segmentation (presented in \tabref{tab:sb16rcnn} and \tabref{tab:maskrcnn}) which highlights the beneficial impact of using \net~when only data from the target task is available.

\begin{table}
\footnotesize
\centering
\caption{Evaluation of noise dataset on object detection and instance segmentation on Sugar Beets 2016. All metrics are reported in [$\%$] and averaged over three runs.}
\label{tab:noisedataset}
\setlength\tabcolsep{3.7pt}
 \begin{tabular}{l|ccc|ccc}
 \toprule
  & \multicolumn{3}{c|}{Object Detection} & \multicolumn{3}{c}{Instance Segmentation} \\
 \textbf{Noise dataset} & \textbf{AP\textsubscript{crop}} & \textbf{AP\textsubscript{weed}} &\textbf{mAP} & \textbf{AP\textsubscript{crop}} & \textbf{AP\textsubscript{weed}} &\textbf{mAP} \\
 \midrule
SB16 \citeS{S-Chebrolu2017AgriculturalRD} & 56.19 & 21.86 & 39.03 & 50.88 & 15.19 & 33.04 \\
Synthetic SB16 \citeS{fppbnRAS2021} & 67.10 & 21.98 & 44.54 & 54.45 & \textbf{15.94} &	35.20 \\
FP22 (ours) & \textbf{69.46} & \textbf{22.25} & \textbf{45.86} & \textbf{54.96} & 15.48 & \textbf{35.22} \\ 
OPPD \citeS{pheno2020} & 63.30 & 22.16 & 42.73 & 51.06 & 15.62 & 33.34 \\
MinneApple \citeS{hani2019minneapple} & 61.01	& 21.75	& 41.38	& 52.69	& 15.35	& 34.02 \\
\midrule
Cityscapes \citeS{Cordts2016TheCD} & 66.39	& 21.76	& 44.07	& 54.85	& 15.32	& 35.09 \\
Random Pixels & 63.99 & 21.96 & 42.97	& 53.46	& 15.56	& 34.51 \\
\bottomrule
\end{tabular}
\vspace{-0.2cm}
\end{table}

\begin{figure*}
\centering
\footnotesize
\setlength{\tabcolsep}{0.05cm}
{
\renewcommand{\arraystretch}{0.2}
\newcolumntype{M}[1]{>{\centering\arraybackslash}m{#1}}
\begin{tabular}{M{2.6cm}M{2.6cm}M{2.6cm}M{2.6cm}M{2.6cm}M{2.6cm}}
\includegraphics[width=\linewidth, frame]{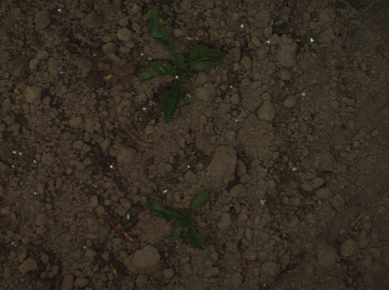} & 
\includegraphics[width=\linewidth, frame]{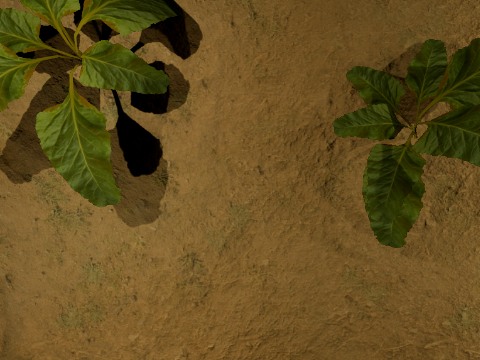} & 
\includegraphics[width=\linewidth, frame]{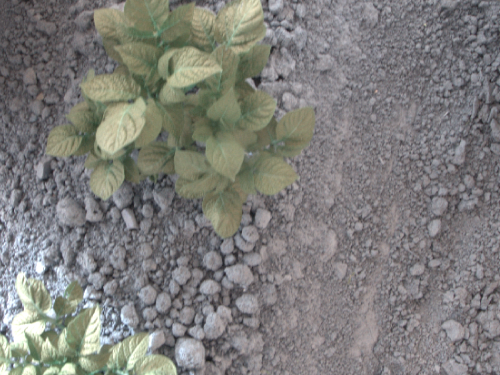} &
\includegraphics[width=\linewidth, frame]{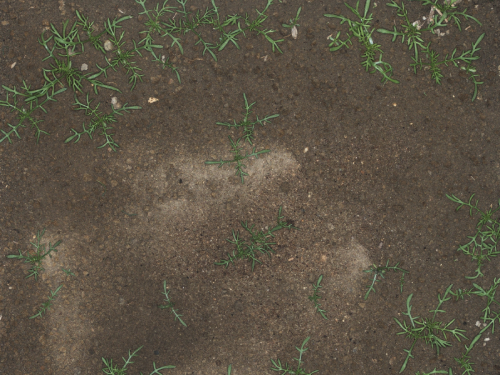} & 
\includegraphics[width=\linewidth, frame]{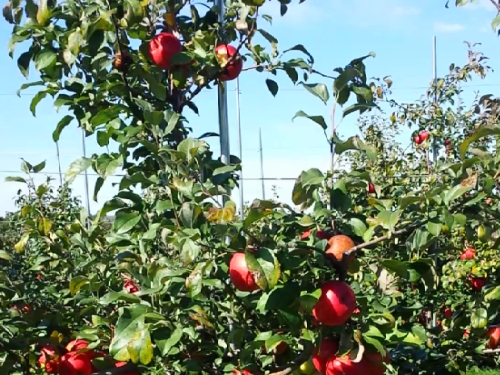}
& \includegraphics[width=\linewidth, frame]{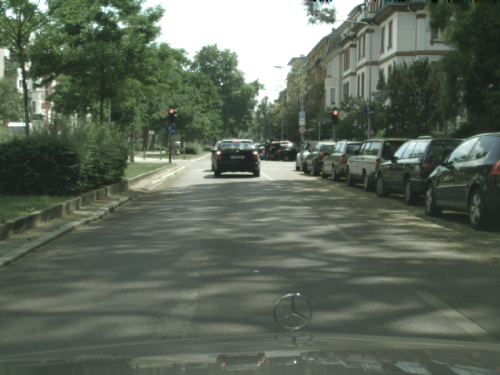} \\ \\ 
(i) SB16 \citeS{S-Chebrolu2017AgriculturalRD} & (ii) Synthetic SB16 \citeS{fppbnRAS2021} & (iii) FP22 (ours) & (iv) OPPD \citeS{pheno2020} & (v) MinneApple \citeS{hani2019minneapple} & (vi) Cityscapes \citeS{Cordts2016TheCD}\\
\end{tabular}
}
\caption{Sample images of applied noise datasets.}
\label{fig:samples}
\vspace{-0.3cm}
\end{figure*}

\subsection{Pretraining Task Transfer} \label{sec:ssl-transfer}
We demonstrate the versatility of our model by (i) pretraining it on all three tasks and finetuning it on object detection and (ii) pretraining it on semantic segmentation and finetuning it on object detection. We compare these results to those obtained when the network is both pretrained and finetuned on object detection. For (ii), we use the pretrained weights from the semantic segmentation network used for experiments outlined in \tabref{tab:semseg}, and finetune them on the object detection task. Consequently, we only transfer the pretrained ResNet50-backbone in this setting. We report the results of these experiments in \tabref{tab:transfer}. Pretraining on semantic segmentation outperforms the best baseline, i.e., ImageNet pretraining, on the AP\textsubscript{crop} and mAP metrics which highlights the learning of semantically meaningful encodings of \net. This experiment highlights the fact that the finetuning procedure is agnostic to the pretraining task and only relies on strong semantic knowledge learned by the network during the pretraining phase. However, we note that the finetuned network yields comparable, yet slightly better results, when the pretraining and target tasks are identical. We thus recommend using the same task for both training phases for achieving optimal performance. However, pretraining on all tasks simultaneously does not result in favorable performance on the weed class. We hypothesize that this disparity is largely a result of sub-optimal weighting of the different loss terms, and further ablative experiments are needed to determine the suitable weighting parameters.

\begin{table}
\footnotesize
\centering
\caption{Evaluation of object detection on the Fraunhofer Potato 2022 dataset. All metrics are reported in [$\%$] and averaged over three runs.}
\label{tab:transfer}
\setlength\tabcolsep{3.7pt}
 \begin{tabular}{l|l|cc|c}
 \toprule
 \textbf{Pretraining Task} & \textbf{Target Task} & \textbf{AP\textsubscript{crop}} & \textbf{AP\textsubscript{weed}} &\textbf{mAP} \\
 \midrule
 Supervised (IN) & Object Det. & 56.77 & \textbf{35.07} & 45.92 \\ 
 \midrule
All & Object Det. & 59.53 & 31.87 & 45.70 \\
Semantic Seg. & Object Det. & 59.49 & 33.75 & 46.62 \\
Object Det. & Object Det. & \textbf{60.85} & 33.24 & \textbf{47.05} \\
\bottomrule
\end{tabular}
\vspace{-0.2cm}
\end{table}

\section{Implementation Details} \label{sec:impl}

\subsection{Agri. BarlowTwins} \label{sec:ssl-agri-barlow}
We use a custom implementation of~\citeS{S-roggiolani2023icra-odsp} where we combine the hyperparameters provided in BarlowTwins~\citeS{S-Zbontar2021BarlowTS} with the domain-adapted augmentations proposed in~\citeS{S-roggiolani2023icra-odsp} as the original implementation of~\citeS{S-roggiolani2023icra-odsp} does not yield representative results in our setting. Specifically, we employ the LARS optimizer, train on batches of size $256$, and scale the learning rates of weights and biases to $0.025$ and $0.0006$, respectively. Further, we extend the proposed augmentations in~\citeS{S-roggiolani2023icra-odsp} with random cropping and image normalization to ensure fairness with the other self-supervised baselines. We compare the performances achieved by the different implementations for object detection on our Fraunhofer Potato dataset in~\tabref{tab:bt}.

\begin{table}
\footnotesize
\centering
\caption{Comparison of BarlowTwins hyperparameter settings on the Fraunhofer Potato 2022 dataset. All metrics are reported in [$\%$] and averaged over three runs.}
\label{tab:bt}
\setlength\tabcolsep{3.7pt}
 \begin{tabular}{l|cc|c}
 \toprule
 \textbf{Pretraining Task} & \textbf{AP\textsubscript{crop}} & \textbf{AP\textsubscript{weed}} &\textbf{mAP} \\
 \midrule
Roggiolani et al.~\citeS{S-roggiolani2023icra-odsp} & 44.44 & 26.34 & 35.39 \\
BarlowTwins~\citeS{S-Zbontar2021BarlowTS} & \textbf{58.16} & 32.99 & 45.58 \\
Our implementation of Roggiolani et al.~\citeS{S-Zbontar2021BarlowTS, S-roggiolani2023icra-odsp} & 57.70 & \textbf{33.64} & \textbf{45.67} \\
\bottomrule
\end{tabular}
\vspace{-0.2cm}
\end{table}

\newpage
{\footnotesize
\bibliographystyleS{IEEEtran}
\bibliographyS{references}
}

\end{document}